\documentclass[10pt,twocolumn,letterpaper]{article}

\usepackage[pagenumbers]{cvpr} 

\usepackage{graphicx}
\usepackage{amsmath}
\usepackage{amssymb}
\usepackage{booktabs}
\usepackage{appendix}
\usepackage{tabularx}
\usepackage{threeparttable}
\usepackage{subcaption}
\usepackage{multirow}

\usepackage[pagebackref,breaklinks,colorlinks]{hyperref}

\usepackage[capitalize]{cleveref}
\crefname{section}{Sec.}{Secs.}
\Crefname{section}{Section}{Sections}
\Crefname{table}{Table}{Tables}
\crefname{table}{Tab.}{Tabs.}
\newcommand{\tocite}[1]{\textcolor{red}{[TO CITE]}}

\def\cvprPaperID{*****} 
\def\confName{CVPR}
\def\confYear{2023}

\newcommand\nonumfootnote[1]{%
\begingroup%
    \renewcommand\thefootnote{}\footnote{\hspace{-3.7pt}#1}%
    \addtocounter{footnote}{-1}%
\endgroup%
}

\begin{document}

\title{
Improved Real-time Image Smoothing with Weak Structures Preserved and High-contrast Details Removed
}

\author{
    \vspace{2 mm}
    Shengchun Wang$^{1,2}$\:
   	Wencheng Wang$^{1,2*}$\:
    Fei Hou$^{1,2}$ \\
    $^1$The State Key Laboratory of Computer Science, Institute of Software, \\
     Chinese Academy of Sciences\:\\
    $^2$School of Computer Science and Technology, University of Chinese Academy of Sciences \\
    \textbf{\{wangsc,whn,houfei\}@ios.ac.cn}
}

\twocolumn[{
\renewcommand\twocolumn[1][]{#1}
\maketitle
\begin{center} 
    \centering
    \setcounter{figure}{0}
    \captionsetup{type=figure}
	\subfloat[\small Input]{
		\begin{minipage}[b]{0.155\linewidth}
			\includegraphics[width=1\linewidth]{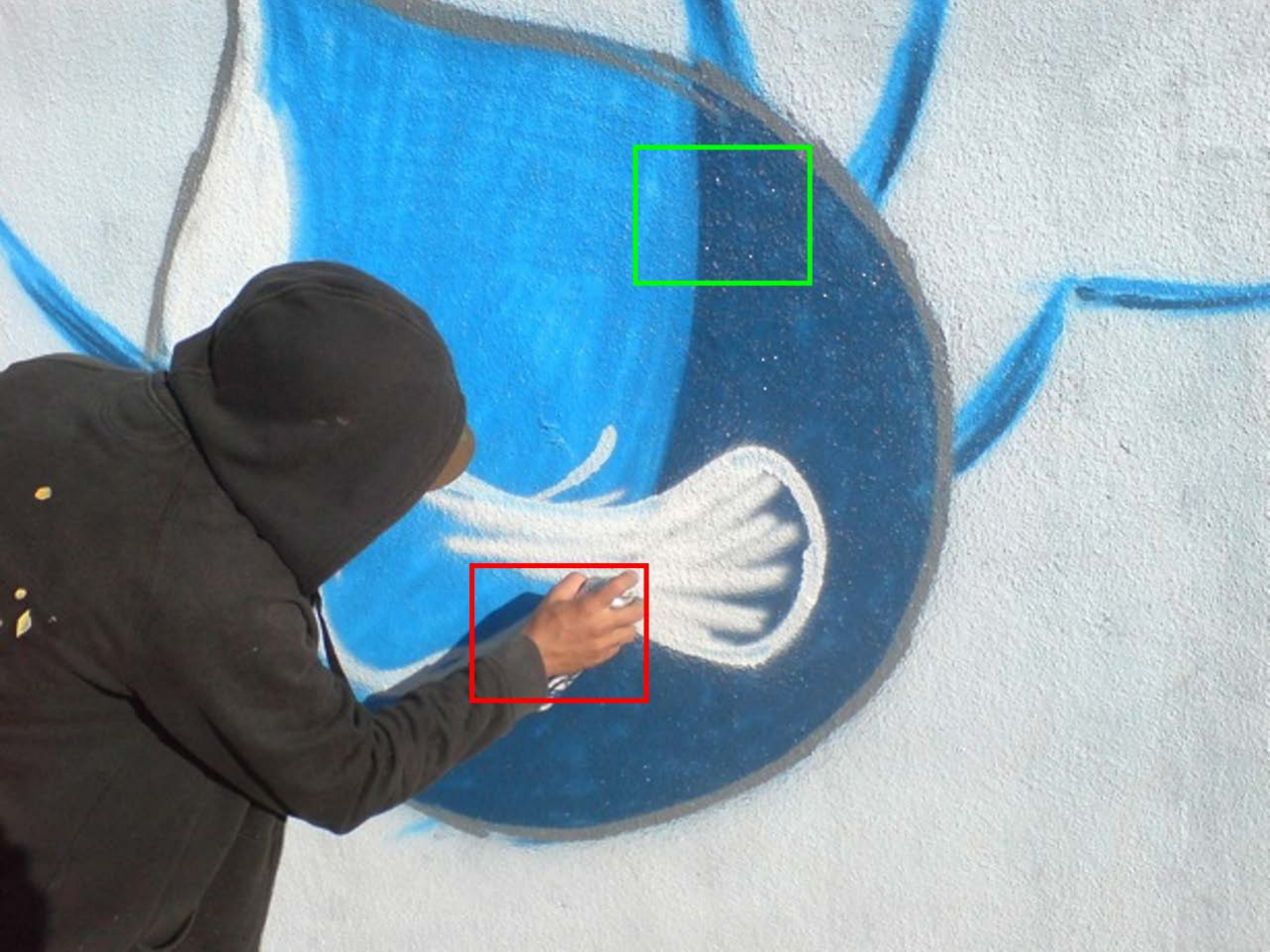}
			\begin{minipage}[b]{0.48\linewidth}
				\includegraphics[width=1\linewidth]{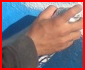}
			\end{minipage}
			\begin{minipage}[b]{0.48\linewidth}
				\includegraphics[width=1\linewidth]{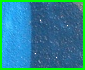}
			\end{minipage}
		\end{minipage}
		\label{fig1a}}	
	\subfloat[\small GFES]{
		\begin{minipage}[b]{0.155\linewidth}
			\includegraphics[width=1\linewidth]{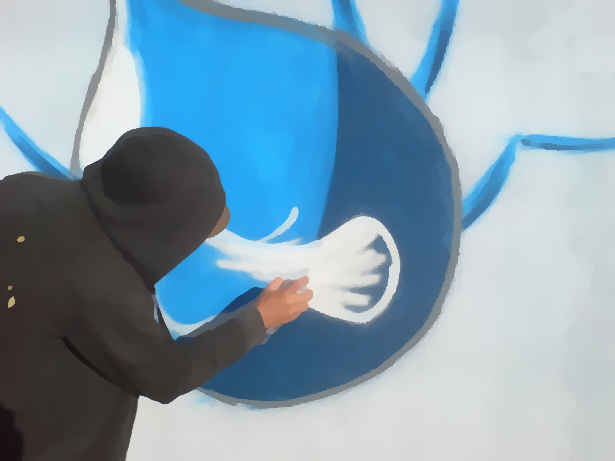}
			\begin{minipage}[b]{0.48\linewidth}
				\includegraphics[width=1\linewidth]{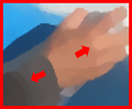}
			\end{minipage}
			\begin{minipage}[b]{0.48\linewidth}
				\includegraphics[width=1\linewidth]{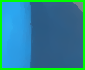}
			\end{minipage}
	\end{minipage}}
	\subfloat[\small DeepFSPIS]{
		\begin{minipage}[b]{0.155\linewidth}
			\includegraphics[width=1\linewidth]{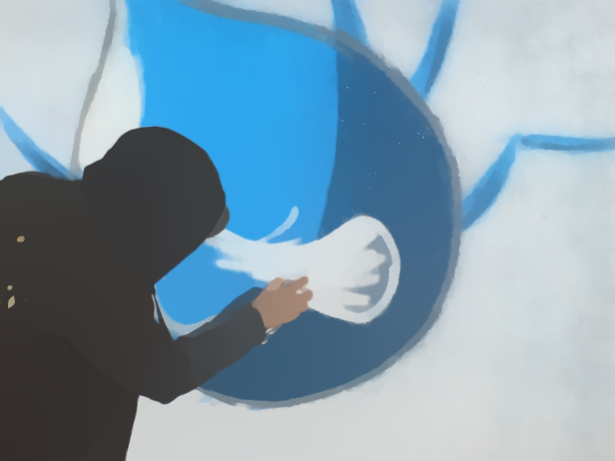}
			\begin{minipage}[b]{0.48\linewidth}
				\includegraphics[width=1\linewidth]{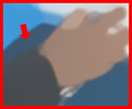}
			\end{minipage}
			\begin{minipage}[b]{0.48\linewidth}
				\includegraphics[width=1\linewidth]{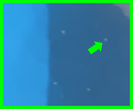}
			\end{minipage}
		\end{minipage}
		\label{fig1b}}
	\subfloat[\small CSGIS-Net]{
		\begin{minipage}[b]{0.155\linewidth}
			\includegraphics[width=1\linewidth]{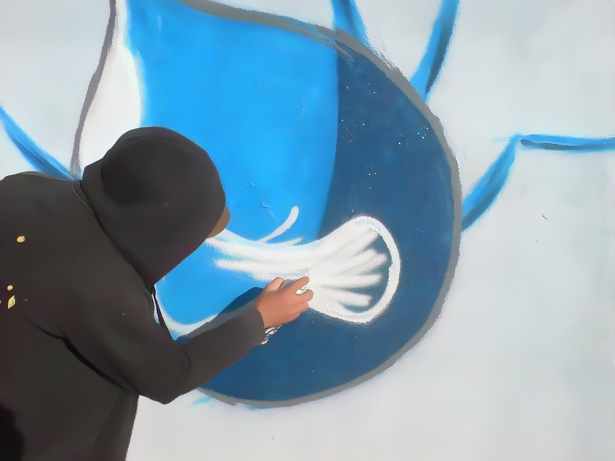}
			\begin{minipage}[b]{0.48\linewidth}
				\includegraphics[width=1\linewidth]{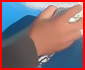}
			\end{minipage}
			\begin{minipage}[b]{0.48\linewidth}
				\includegraphics[width=1\linewidth]{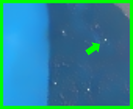}
			\end{minipage}
	\end{minipage}}
	\subfloat[\small ILS]{
		\begin{minipage}[b]{0.155\linewidth}
			\includegraphics[width=1\linewidth]{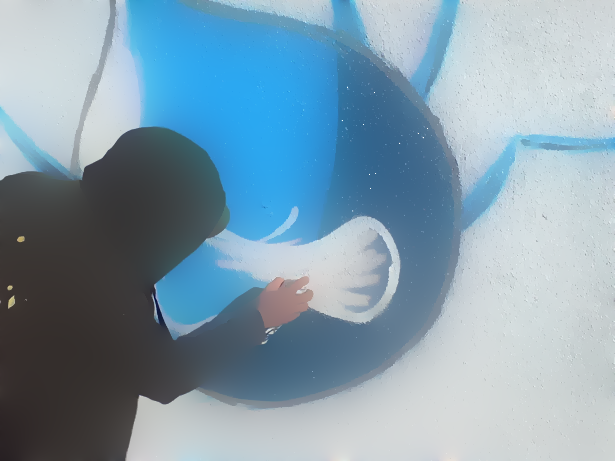}
			\begin{minipage}[b]{0.48\linewidth}
				\includegraphics[width=1\linewidth]{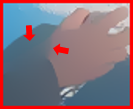}
			\end{minipage}
			\begin{minipage}[b]{0.48\linewidth}
				\includegraphics[width=1\linewidth]{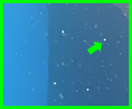}
			\end{minipage}
		\end{minipage}
		\label{fig1c}}
	\subfloat[\small Ours]{
		\begin{minipage}[b]{0.155\linewidth}
			\includegraphics[width=1\linewidth]{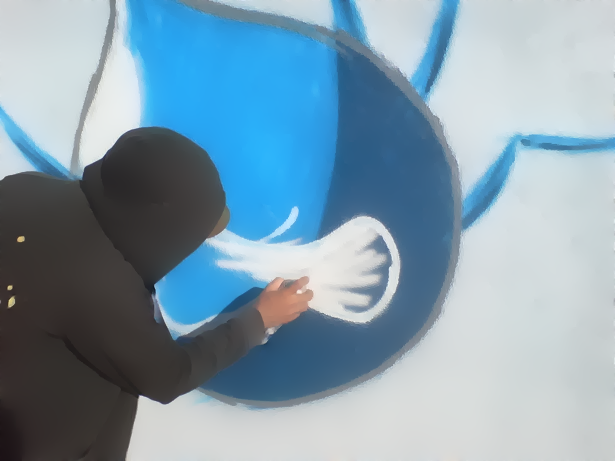}
			\begin{minipage}[b]{0.48\linewidth}
				\includegraphics[width=1\linewidth]{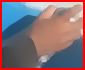}
			\end{minipage}
			\begin{minipage}[b]{0.48\linewidth}
				\includegraphics[width=1\linewidth]{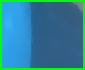}
			\end{minipage}
	\end{minipage}}
    \caption{Comparison between some latest methods and our improved method. For the input (a), GFES~\cite{liu2021generalized} produces block effects, as shown in the red box in (b), DeepFSPIS~\cite{li2022deep}, CSGIS-Net \cite{wang2022contrastive} and ILS \cite{liu2020real} fail to remove some high-contrast details in white dots, as shown in the green boxes in (c), (d) and (e), and DeepFSPIS and ILS have weak structures smoothed out, as shown in the enlarged red boxes in (c) and (e). As for our result (f), it is in high quality without high-contrast details and has weak structures well preserved. Compared with ILS, which is the fastest no-learning method to our knowledge, we took 2 iterations while ILS took 10 iterations to produce these results. (Zoom in for a better view.)
    }
	\label{fig1}
\end{center} 
}]

\begin{abstract}
\nonumfootnote{$^*$Corresponding Author.}
Image smoothing is by reducing pixel-wise gradients to smooth out details. As existing methods always rely on gradients to determine smoothing manners, it is difficult to distinguish structures and details to handle distinctively due to the overlapped ranges of gradients for structures and details. Thus, it is still challenging to achieve high-quality results, especially on preserving weak structures and removing high-contrast details. In this paper, we address this challenge by improving the real-time optimization-based method via iterative least squares (called {\itshape ILS}). We observe that 1) ILS uses gradients as the independent variable in its penalty function for determining smoothing manners, and 2) the framework of ILS can still work for image smoothing when we use some values instead of gradients in the penalty function. Thus, corresponding to the properties of pixels on structures or not, we compute some values to use in the penalty function to determine smoothing manners, and so we can handle structures and details distinctively, no matter whether their gradients are high or low. As a result, we can conveniently remove high-contrast details while preserving weak structures. Moreover, such values can be adjusted to accelerate optimization computation, so that we can use fewer iterations than the original ILS method for efficiency. This also reduces the changes onto structures to help structure preservation. Experimental results show our advantages over existing methods on efficiency and quality.

\end{abstract}

\section{Introduction}
\label{sec:intro}
Image smoothing is to smooth out details while preserving structures to concisely present the contents in the images, by which many subsequent image processing applications can be facilitated, like saliency detection \cite{perazzi2012saliency}, image abstraction \cite{kang2008flow}, pencil sketching \cite{xu2011image}, and detail enhancement \cite{fattal2007multiscale}. Till now, a large number of image smoothing methods have been proposed. They are either filtering-based methods~\cite{he2010guided,cho2014bilateral,jeon2016scale,yin2019side,sun2022edge} or optimization-based methods~\cite{farbman2008edge,xu2012structure,zhang2020erasing,liu2021generalized}. Though they take different strategies for smoothing out details, they all rely on gradients to determine smoothing manners and so difficult to distinguish structures and details to handle distinctively, because the ranges of gradients for structures and details may be overlapped. As a result, it is still very challenging to achieve high-quality results, especially on preserving weak structures while removing high-contrast details. For example, the filtering-based methods calculate the output pixel intensity as a weighted average of input pixel intensities inside a window. Their results are dependent on the windows. The smaller windows help for preserving structures while the larger windows for smoothing out details. As the windows are determined by some measurement on gradients, their structure-preserving abilities and their smoothing abilities are difficult to be well balanced. As discussed in \cite{xu2020pixel}, these methods tend to produce artifacts like halos and gradient reversals. As for optimization-based methods, they formulate image smoothing as an optimization problem to solve globally, generally in an iteration manner. They can achieve superior performance over the filtering-based methods in avoiding artifacts. Unfortunately, they are by iteratively reducing pixelwise gradients for smoothing, so that it is in the dilemma of smoothing out high-contrast details and preserving weak structures, which have low contrasts. Using fewer iterations, the high-contrast details cannot be smoothed. If many more iterations are used, weak structures would be smoothed out. To our knowledge, some learning methods have been proposed recently for image smoothing \cite{chen2017fast,lu2018deep,zhu2019benchmark,wang2022contrastive,li2022deep}. With the trained networks, they can fast output the results. However, they always need an expensive training process and the quality of their results suffers from the training data, which are always produced by existing non-learning methods for image smoothing or using human-made images. Thus, their potentials are limited, especially in handling real-world images.

In this paper, we address this challenge of high-quality image smoothing by determining smoothing manner for pixels via their properties on structures or not, no matter whether they have high or low gradients. In this way, it is convenient to preserve weak structures while removing out high-contrast details in image smoothing. Our proposed method is based on improving the ILS method in \cite{liu2020real}, which is optimization-based and solves an objective function through iterative least squares, and so-called {\it ILS}. It exploits frequency computation for solving the objective function and so able to perform real-time image smoothing. As an optimization-based method, it is ineffective to preserve weak structures and remove high-contrast details, as discussed in the above paragraph. We observe that ILS prefers to smooth out the pixels with lower gradients while not for the pixels with higher gradients, and this is much dependent on the computation of its penalty function, whose independent variable is the pixelwise gradient. This motivates us that the values of the penalty function take much effects on determining smoothing manners. Thus, we modify the computation of the penalty function by the properties of pixels on structures or not, allowing the pixels of details to have lower penalty function values and the pixels of the structures to have higher penalty function values, by which the smoothing would be dependent on whether they should be smoothed or preserved, no matter whether these pixels have high or low gradients. By an analysis, with such a modification of the penalty function, the framework of the ILS method can be still used for image smoothing. Thus, we can more effectively smooth out the details with high contrasts, and due to this, we can use fewer iterations than the original ILS method to obtain quality results, by which efficiency can be promoted and structures can be more effectively preserved due to the reduced changes onto structures, e.g., avoiding the halos and intensity shift. All these will be discussed in detail in Section~\ref{sec:ExILS} and \ref{sec:weight}. Experimental results show that we can obtain better results than state-of-the-art methods, especially on removing high-contrast details and preserving weak structures, while we can reduce the iterations considerably in comparison with the original ILS method for acceleration, as illustrated in Figure \ref{fig1}.

\section{Related Works}
\label{sec:relateWork}

\subsection{Filtering-based Methods}
\label{subsec:Filterbased}
Filtering-based methods calculate the output pixel intensity as a weighted average of input pixel intensities inside a window, such as the popular bilateral filter \cite{tomasi1998bilateral}. For improvement, some methods try to enhance weight computation, like using histograms \cite{felsberg2005channel,van2001local,kass2010smoothed} or prioritizing spatial scales \cite{subr2009edge}, while some try to promote the efficiency, including fast bilateral filter \cite{barron2016fast},  adaptive manifolds for real-time high-dimensional filtering \cite{gastal2012adaptive} and fast high-dimensional filtering using the permutohedral lattice \cite{adams2010fast}. Considering the results are much determined by the scopes defined by the windows, many methods try to improve window determination, e.g. using non-local windows in the tree filter \cite{bao2013tree} and the graph-based filter \cite{zhang2015segment}, using small windows near edges and using large windows inside the regions far away from edges \cite{jeon2016scale,song2019scale,sun2022edge} and using edge-aware windows \cite{zang2015guided,xu2018improved,xu2019structure}. Some methods also investigate to employ priors for improvements, like guided image filtering \cite{he2012guided,li2014weighted,sun2019weighted}. These methods are generally very efficient, but they are prone to produce artifacts in their results, as discussed in \cite{xu2020pixel,li2022deep}.

\subsection{Optimization-based Methods }
\label{subsec:optimibsaed}
Optimization-based methods~\cite{xu2012structure,zhang2020erasing,zhou2020structure,liu2021generalized} take image smoothing as an optimization problem to solve, by which details are smoothed out while structures are preserved. As they globally consider the characteristics of pixels and may embed the prior of the output image in the optimization procedure to guide image smoothing, they are better in avoiding artifacts than filtering-based methods, e.g. the methods via total variation smoothing \cite{rudin1992nonlinear}, weighted least square filter \cite{farbman2008edge} and $L_{0}$ smoothing \cite{xu2011image}. Considering textural details may contain significant local gradients, some methods are particularly studied to improve texture filtering \cite{xu2012structure,ham2015robust}. As global optimization based methods are computationally expensive, many methods are proposed to speed up optimization computation, like using preconditioned techniques \cite{afonso2010fast,krishnan2013efficient}. With regard to this, Liu et al. \cite{liu2020real} proposed a method using iterative least squares for fast solving the optimization problem via fast Fourier transforms and inverse fast Fourier transforms. It can perform image smoothing in real time while preserving salient edges. In general, these methods achieve smoothing results by imposing certain penalty on image gradients, so that they are often difficult to smooth out the details with high-contrasts while preserving weak structures, as discussed in Section \ref{sec:intro}. This prevents them from obtaining high quality results.

\subsection{Learning-based Methods}
\label{subsec:learnbased}
Recently, learning techniques have been studied to promote image smoothing. Some methods use deep neural networks (DNN) architectures as a solver to solve the objective function or predict parameters \cite{fan2018image,zhou2020structure,kim2021deformable,shi2021unsharp}. As they need to train the model separately for different inputs, they are not easy to use. Some methods design novel network architectures as the image smoothing models to directly predict smoothing results by ground truth datasets \cite{xu2015deep,lu2018deep,zhu2018saliency,zhu2019benchmark,gao2020semi,xu2020pixel,feng2021easy2hard,wang2022contrastive}. Though they are very convenient to output the results after their networks are learned, their results are prevented in quality by their training data. This is because the training data are always produced by existing non-learning methods, whose shortcomings would be transferred to learning methods, as discussed in ~\cite{lu2018deep,zhu2019benchmark}. Though Fan, et al. \cite{fan2018image} proposed an unsupervised learning method, it relies on the detected textures or structures to optimize the objective functions during training. Since high quality textures or structures are difficult to obtain, the potentials of this method are limited. Till now, it is still difficult for learning methods to produce high quality results.
\section{Determining Smoothing Manners by properties of pixels on structures}
\label{sec:ExILS}
Our method is by modifying the penalty function of ILS to decouple gradients from determining smoothing manners and then employ the framework of ILS for image smoothing. In the following subsections, we first review ILS and then discuss our improvements.

\subsection{ILS Method Review}
\label{subsec:ILS}

The ILS method was proposed in~\cite{liu2020real}, which is by minimizing the following energy function:
\begin{equation}
	\label{eq1}
	E(u,f) = \sum_{s} \left( (u_{s} - f_{s})^{2} + \lambda \sum_{* \in {x,y}} \phi_p(\nabla u_{*,s}) \right),
\end{equation}
where $f$ is the input image, $u$ is the smoothed output image, $s$ denotes the pixel position and $\nabla u_{*}$ $(* \in {x,y})$ represents the gradient of $u$ along $x$-axis/$y$-axis. Here, the gradients are computed by the standard finite difference $[1, -1]$ and $[1, -1]^{\top}$ along $x$-axis and $y$-axis, respectively. The penalty function $\phi_p(\cdot)$ is defined as:
\begin{equation}
	\label{eq2}
	\phi_p(d)=\left(d^2+\epsilon\right)^{\frac{p}{2}}
\end{equation}
where $d$ is the gradient value, $\epsilon$ is a small constant and fixed $\epsilon = 0.0001$ in their tests. The norm power $p$ is usually set as $0 < p \leq 1$ for edge-preserving smoothing and fixed $p = 0.8$ in our tests, as suggested in \cite{liu2020real}.

In \cite{liu2020real}, it is demonstrated that Eq.~(\ref{eq1}) can be solved by updating $u$ iteratively and the value of $u$ in each iteration can be obtained as:

\begin{equation}
	\label{eq3}
	\begin{aligned}
		\begin{split}
			&u^{n+1}= \underset{u}{\arg \min } \sum_s\Biggl(\left(u_s-f_s\right)^2+ \Biggr.\\
			&\Biggl. \lambda \sum_{* \in\{x, y\}} \frac{1}{2}\left(\sqrt{c} \nabla u_{*, s}-\frac{1}{\sqrt{c}} \mu_{*, s}^n \right)^2 \Biggr)
		\end{split}	
	\end{aligned}
\end{equation}
where the constant $c=p \epsilon^{\frac{p}{2}-1}>0$. The value of $\mu_{*, s}^n$ in each iteration is computed as:
\begin{equation}
	\begin{aligned}
		\mu_{*, s}^n & =c \nabla u_{*, s}^n-\phi_p^{\prime}\left(\nabla u_{*, s}^n\right) \\
		& =c \nabla u_{*, s}^n-p \nabla u_{*, s}\left(\left(\nabla u_{*, s}^n\right)^2+\epsilon\right)^{\frac{p}{2}-1}, * \in\{x, y\}
	\end{aligned}
\end{equation}
where $\phi_p^{\prime}$ is the derivative of $\phi_p$. Because each iteration in Eq.~(\ref{eq3}) is a least square (LS) problem, this method is called {\it ILS}.

As Eq.~(\ref{eq3}) can be solved efficiently with the help of fast Fourier transform (FFT) and inverse fast Fourier transform (IFFT), according to the discussion in \cite{liu2020real}, the ILS method is very efficient for image smoothing. Moreover, it is observed that using a few iterations of Eq.~(\ref{eq3}), it is able to achieve most of the energy decrease, so that the edges will not be influenced very much, helpful for edge-preserving in image smoothing, which is also benefited from the suitable parameter settings, as discussed in \cite{liu2020real}.

\subsection{Improvements }
\label{sec:improve}

Though ILS is efficient and able to preserve edges, it is unable to smooth out high-contrast details and would smooth out weak structures (represented by edges), as discussed in \cite{liu2020real}. Here, we present improvements to address these shortcomings. 

\begin{figure}[htb]
	\centering
	\begin{minipage}{0.35\linewidth}
		\includegraphics[width=\linewidth]{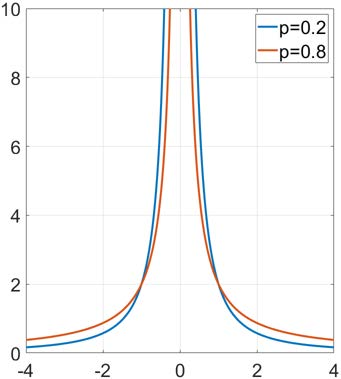}
	\end{minipage}
	\caption{The pixels with lower gradients (horizontal axis) have higher values for edge stopping function (vertical axis), meaning they are smoothed much more, no matter what $p$ is. (The figure is in courtesy of \cite{liu2020real}.)}
	\label{fig2}
\end{figure}

By the plots in Figure~\ref{fig2} about the relationship between gradients and the smoothing effects, represented by the values of edge stopping function, we observe that the penalties are corresponding to gradients monotonously, and ILS tends to smooth out the pixels with lower gradients more preferentially than the pixels with higher gradients. This motivates us that the values of the penalty function for pixels determine how the pixels are smoothed. For the pixels with lower values of the penalty function, they will be smoothed more preferentially, while not for the pixels with higher values of the penalty function. 

Thus, if we have the penalty function computed by some values instead of gradients, called {\it guidance values}, these guidance values would take effects to determine how the pixels are smoothed. With an analogical reasoning like the discussion in the above paragraph, the pixels with lower guidance values would be much more smoothed while not for the pixels with higher guidance values. Therefore, when the pixels are determined in the regions to be much smoothed, they will adjust their intensities to be very smooth no matter whether their gradients are lower or higher. Similarly, for the pixels determined to be little smoothed, they will not be smoothed much and so their gradients would be well kept for helping preservation of their related structures (edges), even the weak structures. 

\begin{figure}[tbp]
	\centering
	\subfloat{
		\begin{minipage}[t]{0.3\linewidth}
			\centering
			\includegraphics[width=1\linewidth]{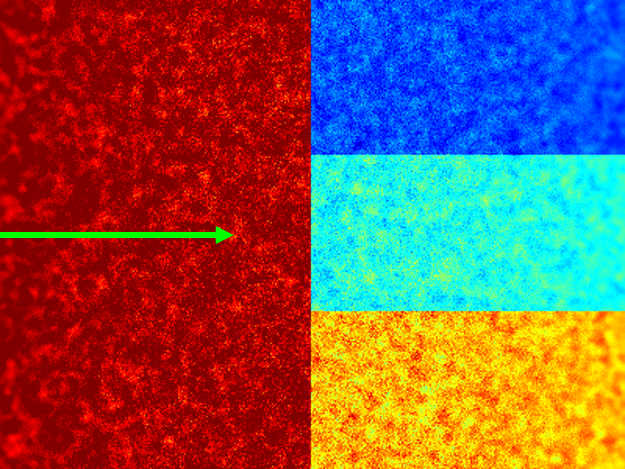}			
		\end{minipage}
	}
	\subfloat{
		\begin{minipage}[t]{0.3\linewidth}
			\centering
			\includegraphics[width=1\linewidth]{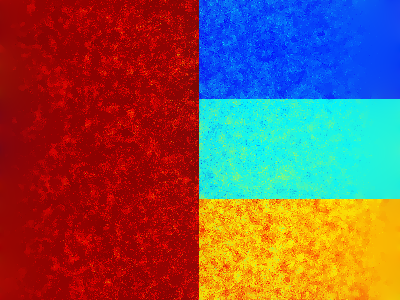}
		\end{minipage}
	}
	\subfloat{
		\begin{minipage}[t]{0.3\linewidth}
			\centering
			\includegraphics[width=1\linewidth]{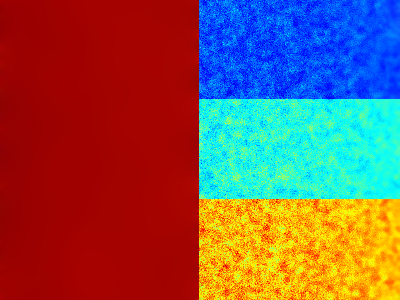}
		\end{minipage}
	}
	\setcounter{subfigure}{0}
	\subfloat[]{
		\begin{minipage}[t]{0.3\linewidth}
			\centering
			\includegraphics[width=1\linewidth]{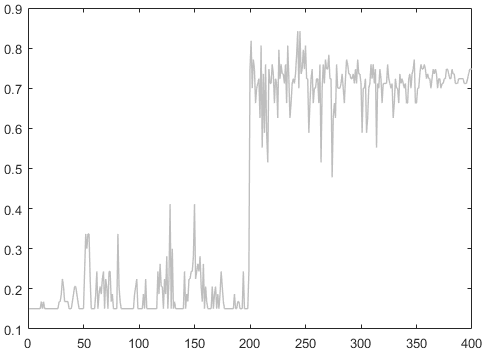}
		\end{minipage}
		\label{fig3a}
	}
	\subfloat[]{
		\begin{minipage}[t]{0.3\linewidth}
			\centering
			\includegraphics[width=1\linewidth]{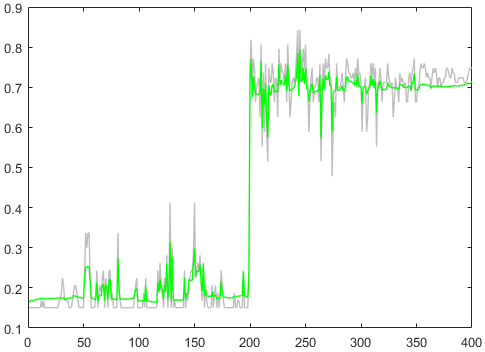}
		\end{minipage}
		\label{fig3b}
	}
	\subfloat[]{
		\begin{minipage}[t]{0.3\linewidth}
			\centering
			\includegraphics[width=1\linewidth]{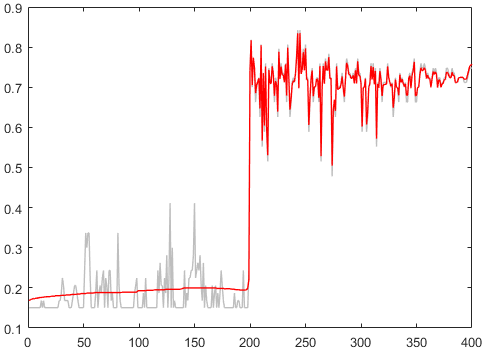}
		\end{minipage}
		\label{fig3c}
	}
	\caption{The red region can be well smoothed with our improved method, where we assign $\omega$ = 0.1 to the pixels of the red region and $\omega$ = 1.0 to the other pixels in computing the penalty function of the ILS method. (a) Input. (b) The result by the original ILS method with 20 iterations. (c) The result of our improved method with 2 iterations. According to the intensities of the pixels on a scanline, marked by the green arrow in the first row, the pixels in the red region still have high gradients in (b) while not in (c).}
	\label{fig3}
\end{figure}

According to Eq.~(\ref{eq1}), the energy function of the ILS method is not only related to the penalty function, but also related to its data term, trying to remain the pixelwise intensities as much as possible for preserving the contents in the image. Considering gradients are related to intensities and image smoothing is to reduce gradients between pixels of non-structures, we have our guidance values computed by weighting gradients. As a result, we have the penalty function modified as follows,

\begin{equation}
	\label{eq4}
	\phi_{p}(id) =(id^{2} + \epsilon )^{\frac{p}{2}}
\end{equation}
\begin{equation}
	\label{eq5}
	id = \omega_{*,s} \times \nabla u_{*,s}
\end{equation}
where $\omega_{*,s}$ is a weight for determining the values of $id$ to correspond to the properties of pixels on structures or not. It can have the pixel with a high gradient to have a very low $id$ value or have the pixel with a low gradient to have a very high $id$ value. In this way, structures can be preserved and details are removed, irrespective of their gradients. As illustrated in Figure \ref{fig3}, with the red pixels given very lower weights, the red region can be smoothed very well, while this cannot be achieved with the original ILS method.

\section{Weight Computation }
\label{sec:weight} 
In Section \ref{sec:ExILS}, we have discussed that the ILS method can be improved to well smooth out high-contrast details and preserve weak structures when we modify the computation of its penalty function. In this section, we discuss how to compute the weights for achieving high quality results.

According to the discussion in Section \ref{sec:ExILS}, in our modification of the penalty function by Eq.~(\ref{eq4}) and Eq.~(\ref{eq5}), it is expected that the guidance $id$ values are very low for the pixels of details and very high for the pixels of structures. Thus, we need first to determine whether the pixels are on structures or not. Then, we compute the weights in [0.0, 1.0] for keeping guidance $id$ values still in [-1.0, 1.0], by which we can satisfy the requirement to use the framework of ILS for image smoothing, that is, the independent variable of the penalty function must have their values in [-1.0, 1.0]. As discussed in Section \ref{sec:ExILS}, the ILS method prefers to smooth out the pixels with lower guidance values. Therefore, for fast smoothing out the details with high contrasts, it is expected the guidance values for the pixels of details are very near 0.0. As a result, our weight computation is by the following steps:

1) The interval gradient $\nabla_{\Omega}I$ for pixel $q$ is computed, trying to distinguish whether it is for details or on structures. Interval gradients are proposed in \cite{lee2017structure} to enhance the distinguishing of textured details from structures  by enlarging the difference between structures and textures in terms of gradient computation, where the gradient at a pixel is not computed by the difference of intensity between its left and right adjacent pixels, but by the difference between the averaged intensity for a range of pixels to its left and that for a range of pixels to its right. We find interval gradients are also very effective to distinguish other details besides texture details, and so take it for detecting structures.

2) A $\gamma(q)$ value in [0.0, 1.0] is computed by the interval gradient $\nabla_{\Omega}I$ for pixel $q$, trying to allow $\gamma(q)$ to have higher values for structure pixels and lower values for details pixels.

3) The weight $\omega(q)$ is computed by $\gamma(q)$, trying to have its value very near 0.0 for the pixels of details, while not for the pixels of structures.

In the following, these computations are discussed.
\begin{itemize}
	\item{{\bf The interval gradient} $(\nabla_{\Omega}I)_{q}$ in~\cite{lee2017structure} is computed by using a local window $\Omega_{q}$ centered at a pixel $q$ as follows,} 
\end{itemize}

\begin{equation}
	\label{eq6}
	(\nabla_{\Omega}I)_{q} = g_{\sigma}^{r}(I_{q}) - g_{\sigma}^{l}(I_{q})
\end{equation}
where $g_{\sigma}^{r}(I_{q})$ and $g_{\sigma}^{l}(I_{q})$ are left and right clipped 1D Gaussian filter functions defined by

\begin{equation}
	\label{eq7}
	g_{\sigma}^{r}(I_{q}) = \frac{1}{k_{r}} \sum_{n \in \Omega(q) }\omega_{\sigma} (n-q-1)I_{n}
\end{equation}

\begin{equation}
	\label{eq8}
	g_{\sigma}^{l}(I_{q}) = \frac{1}{k_{l}} \sum_{n \in \Omega(q) }\omega_{\sigma} (q-n)I_{n}
\end{equation}
where $k_{r}$ and $k_{l}$ are coefficients for normalization, defined as

\begin{equation}
	\label{eq9}
	k_{r} = \sum_{n \in \Omega(q) }\omega_{\sigma} (n-q-1) \ and \ k_{l} = \sum_{n \in \Omega(q) }\omega_{\sigma} (q-n)
\end{equation}
and $\omega_{\sigma}$ is the clipped exponential weighting function with a scale parameter $\sigma$, which is the range of the neighboring pixels for interval gradient computation in the window $\Omega(q)$. $\omega_{\sigma}$ is defined as

\begin{equation}
	\label{eq10}
	\omega_{\sigma}(x)=\left\{
	\begin{aligned}
		& exp(- \frac{x^{2}}{2\sigma^{2}})  & if \ x \geq 0 \\
		& 0   & otherwise
	\end{aligned}
	\right.
\end{equation}

\begin{itemize}
	\item{ {\bf $\gamma(q)$ computation} is by the following equation,}
\end{itemize}

\begin{equation}
	\label{eq11}
	\gamma(q) = min \left( 1.0,\frac{\mid (\nabla_{\Omega}I)_{q} \mid + \epsilon_{s} }{\mid (\nabla I)_{q} \mid + \epsilon_{s}} \right) 
\end{equation}

\begin{equation}
	\label{eq12}
	(\nabla I)_{q} = I_{q+1} - I_{q}
\end{equation}
where $I$ is a 1D discrete signal, $\epsilon_{s}$ is a small constant to prevent numerical instability and we also fix  $\epsilon_{s}$ = 0.0001 in all the experiments. 

In general, $(\nabla_{\Omega}I)_{q}$ has a smaller absolute value than $(\nabla I)_{q}$ for detailed pixels, so that their $\gamma(q)$ will have a value lower than 1.0. As for a pixel of structures, its $(\nabla_{\Omega}I)_{q}$ always has a larger absolute value than its $(\nabla I)_{q}$, so that its $\gamma(q)$ will have a value 1.0, or much near 1.0.

\begin{itemize}
	\item{ {\bf $\omega(q)$ computation} is by the following equation,}
\end{itemize}

\begin{equation}
	\label{eq13}
	\omega(q) = 2\Big(\frac{1}{1+exp\big(- (2\sigma_{s} + 1) * (\gamma(q) - 1)\big)}\Big) 
\end{equation}
where $\sigma_{s}$ controls the sharpness of weight transition from structures to detail regions, which is fixed as $\sigma_{s}$ = $\sigma$ in all our tests.

With Eq.~(\ref{eq13}), when $\gamma(q)$ is 1.0 or much near 1.0, its $\omega(q)$ will be much near 1.0. With $\gamma(q)$ being smaller and smaller to approach 0.0, the denominator of Eq.~(\ref{eq13}) will increase its value more and more rapidly, and so the $\omega(q)$ will approach 0.0 very much.

\begin{figure}[tb]
	\centering
	\subfloat[input]{\includegraphics[width=0.20\linewidth]{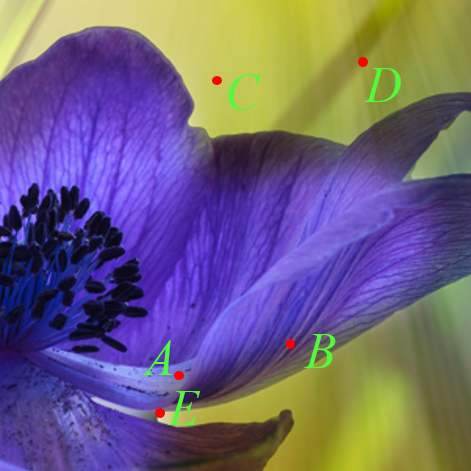}
		\label{fig4a}}
	\hfil
	\subfloat[ILS]{\includegraphics[width=0.20\linewidth]{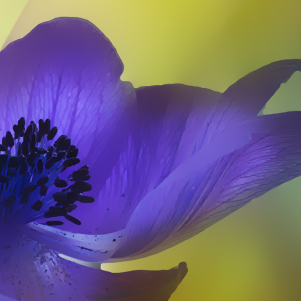}   
		\label{fig4b}}
	\hfil
	\subfloat[Ours]{\includegraphics[width=0.20\linewidth]{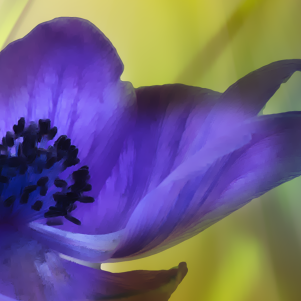}
		\label{fig4c}}
	\subfloat[Our guidance values]{\includegraphics[width=0.33\linewidth]{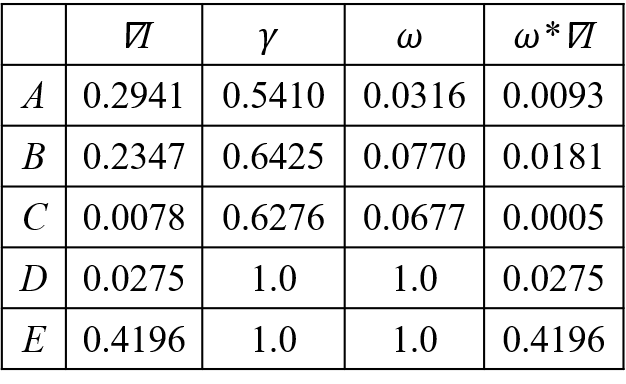}   
		\label{fig4d}}
	\caption{Our weight computation can well correspond to the properties of the pixels on structures or of details. For the pixels of details, $A$, $B$ and $C$, $A$ and $B$ have a little higher gradient $(\nabla I)_{q}$ values while not for $C$. Though their $\gamma(q)$ values are over 0.5, their $\omega(q)$ values are much lower to have their guidance values much near 0.0. As for the pixels on structures, $D$ and $E$, $D$ has a much low gradient $(\nabla I)_{q}$ value on a weak structure while $E$ has a much high gradient $(\nabla I)_{q}$ value. With our weight computation, their $\gamma(q)$ and $\omega(q)$ values are 1.0, so that their guidance values are much higher than those for $A$, $B$ and $C$, by which they are not much smoothed and so their related structures are preserved. Thus, we can produce high quality result (c), where the high-contrast noise point $A$ is well smoothed with $\lambda$ = 0.1 and only 2 iterations in using our improved method. As a comparison in (b), ILS is given a large $\lambda$ = 1.0 value and 4 iterations to have a high smoothing ability, but it cannot yet remove the high-contrast noise point $A$.}
	\label{fig4}
\end{figure}

As illustrated in Figure \ref{fig4}, our weight computation can effectively produce guidance values to well correspond to the properties of the pixels on structures or details, no matter whether the pixels have high or low gradients. Therefore, compared to ILS, our improved method can well preserve weak structures, such as sample $D$, while smoothing out high-contrast details, such as samples $A$ and $B$.

\section{Results and Discussion}
\label{sec:results}

We made tests to compare our method with existing methods, including {\bfseries 1) filtering-based methods} via bilateral texture filtering (BTF) \cite{cho2014bilateral}, interval gradients (IG)~\cite{lee2017structure} and edge guidance filtering for structure extraction (EGF) \cite{sun2022edge}; {\bfseries 2) optimization-based methods} via real-time image smoothing via iterative least squares (ILS) \cite{liu2020real}, erasing appearance preservation (EAP)~\cite{zhang2020erasing}, structure and texture-aware image decomposition via training a neural network (STDN)~\cite{zhou2020structure} and a generalized framework for edge-preserving and structure-preserving image smoothing (GFES)~\cite{liu2021generalized}; and {\bfseries 3) learning-based methods} via deep flexible structure preserving image smoothing (DeepFSPIS)~\cite{li2022deep}, learning to solve the intractable for structure preserving image smoothing  (Easy2Hard)~\cite{feng2021easy2hard} and contrastive semantic-guided image smoothing network (CSGIS-Net)~\cite{wang2022contrastive}. We have our method implemented in MATLAB (R2017b) and download the open codes for all the compared methods, which are provided by the authors, where the non-learning ones are also implemented in MATLAB. We made tests on a personal computer installed with an Intel Core i7-8700 CPU, 48GB RAM, an NVIDIA GeForce GTX 1080Ti GPU with 11 GB memory. The results are always collected with the Windows 10 operating system except the results for EGF, which are collected with Linux operating system (Ubuntu 20.04), and for DeepFSPIS, which are collected from API provided by the author.

\subsection{Parameters }
\label{subsec:parameter}
Our improved method has several parameters. For the parameters $p$, $\epsilon$ and $c$ for ILS computation and $\epsilon_{s}$ for interval gradient computation, they are set as suggested in the references, as they are generally stable for image smoothing. As for our added parameter $\sigma_{s}$ in Eq.~(\ref{eq13}), we only set $\sigma_{s}$ = $\sigma$, and can always obtain good results. 

For high performance, there are three parameters to be well investigated, iteration number $N$, $\lambda$, and $\sigma$. For $N$, when it is larger, the image would be smoothed much more but this will takes much more time, not helpful for preserving edges and tends to produce artifacts like halos and intensity shift. As we can fast smooth out high-contrast details, we can use a few iterations to produce good results. Thus, we used $2 \sim 5$ iterations. 

\begin{figure}[htb]
	\centering
	\subfloat[\small input]{
		\begin{minipage}[t]{0.31\linewidth}
			\centering
			\includegraphics[width=1\linewidth]{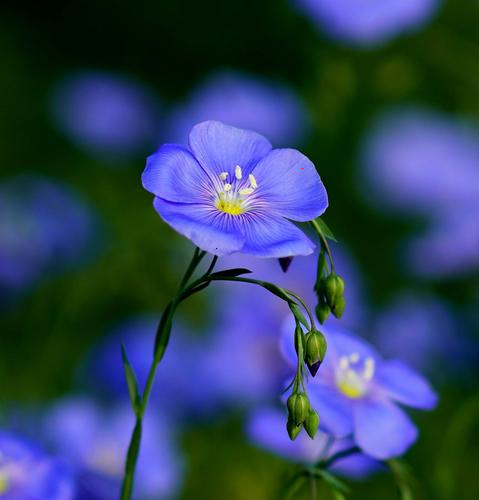}	
		\end{minipage}
	}
	\subfloat[\small $\lambda$ = 0.1, $N$ = 2, $\sigma$ = 2]{
		\begin{minipage}[t]{0.31\linewidth}
			\centering
			\includegraphics[width=1\linewidth]{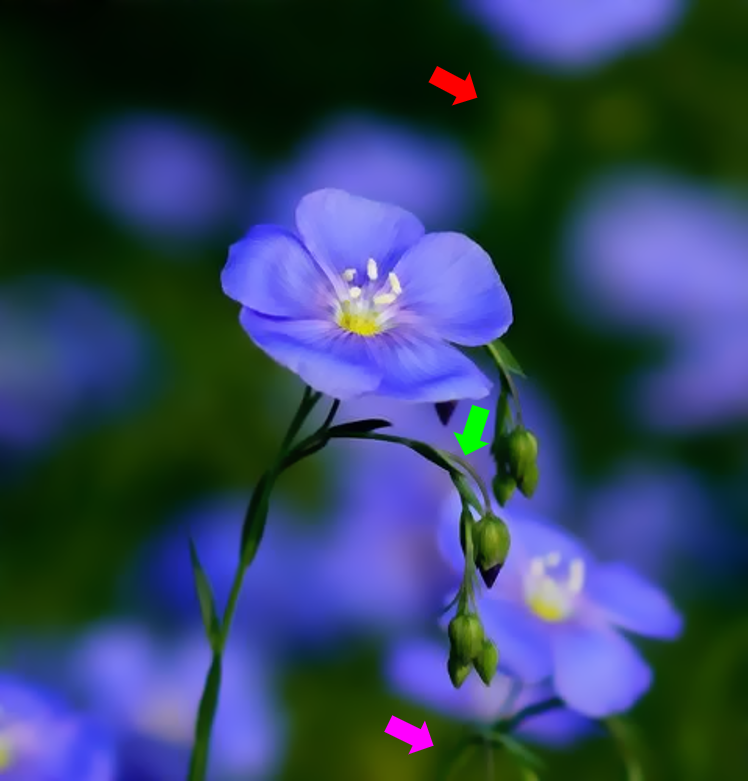}
		\end{minipage}
	}
	\subfloat[\small $\lambda$ = 0.1, $N$ = 2, $\sigma$ = 4]{
		\begin{minipage}[t]{0.31\linewidth}
			\centering
			\includegraphics[width=1\linewidth]{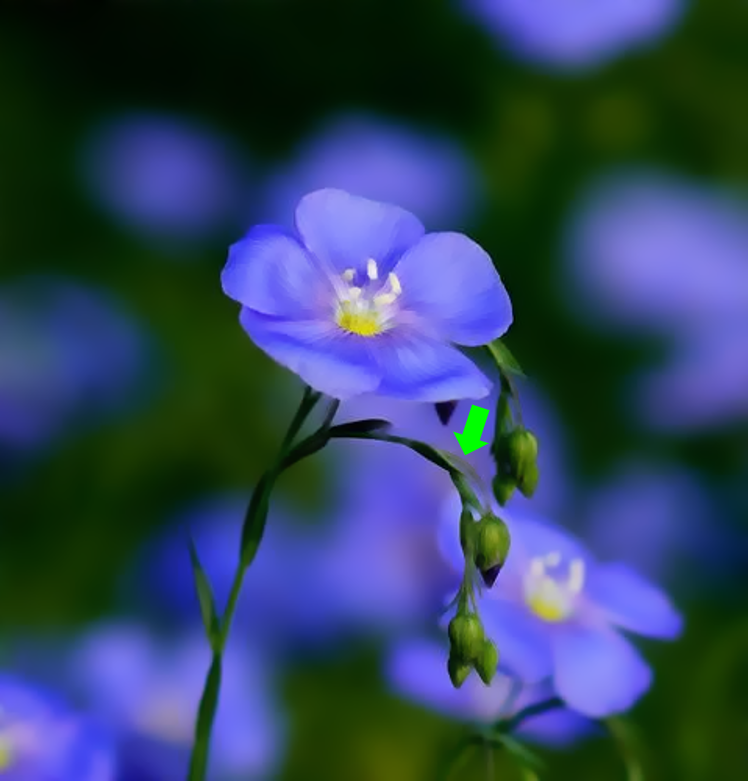}
		\end{minipage}
	}
	\setcounter{subfigure}{3}
	\subfloat[\small $\lambda$ = 0.1, $N$ = 4, $\sigma$ =2]{
		\begin{minipage}[t]{0.31\linewidth}
			\centering
			\includegraphics[width=1\linewidth]{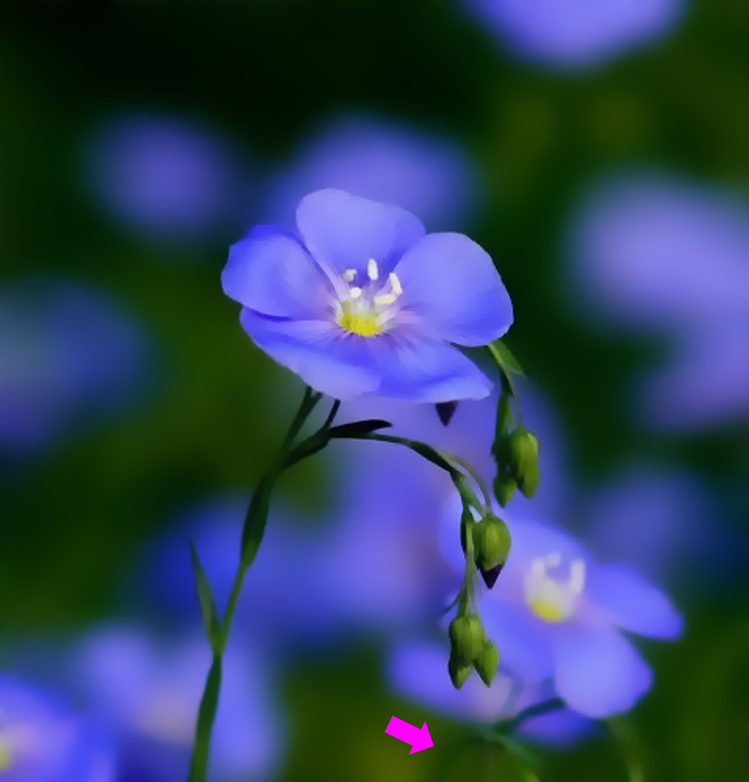}
		\end{minipage}
	}
	\subfloat[\small $\lambda$ = 0.5, $N$ = 2, $\sigma$ =2]{
		\begin{minipage}[t]{0.31\linewidth}
			\centering
			\includegraphics[width=1\linewidth]{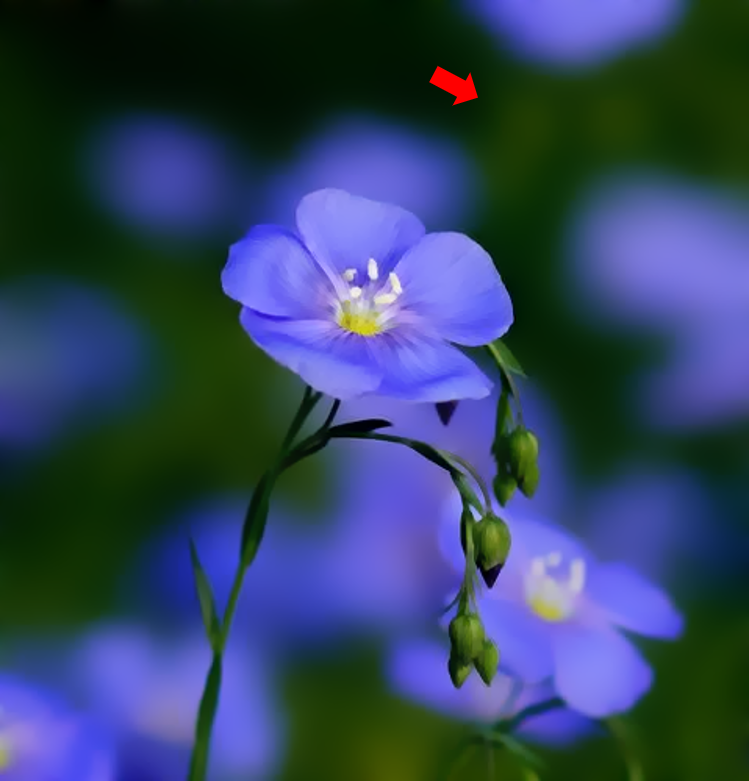}
		\end{minipage}
	}
	\subfloat[\small $\lambda$ = 1, $N$ = 2, $\sigma$ =2]{
		\begin{minipage}[t]{0.31\linewidth}
			\centering
			\includegraphics[width=1\linewidth]{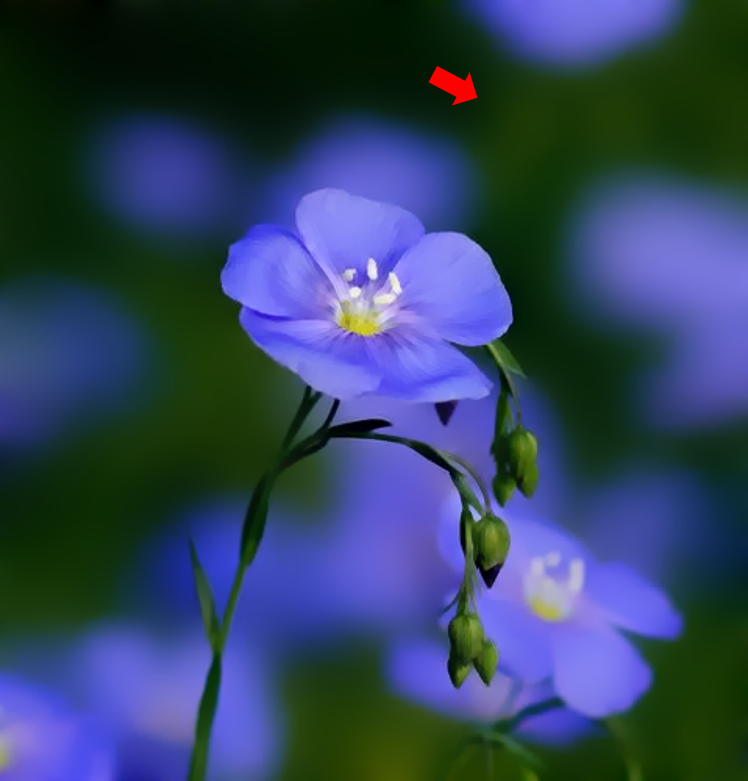}
		\end{minipage}
	}
	\caption{Smoothing results of our method with different settings for $\lambda$, $\sigma$ and $N$. Larger values for $\lambda$, $N$ and $\sigma$ all lead to stronger smoothing on the input image, but this also blurs the structures, especially smoothing out weak structures, as marked by arrows. In applying our method, smaller values for them are preferred.}
	\label{fig5}
\end{figure}

As discussed in \cite{liu2020real}, $\lambda$ controls the smoothing strength and a larger $\lambda$ leads to stronger smoothing. In our tests, it is set $\lambda$ $\in$ [0.1, 1.0]. As for $\sigma$, it controls the scale of details including texture details to be smoothed out, as discussed in \cite{lee2017structure}. In our tests, we always set $\sigma$ $\in$ [2, 5] as they can well produce good results. As illustrated in Figure \ref{fig5}, we can use smaller values for $\lambda$, $N$ and $\sigma$ to achieve good results. 

\begin{figure}
	\centering
	\subfloat{
		\begin{minipage}[b]{0.23\linewidth}
			\includegraphics[width=1\linewidth]{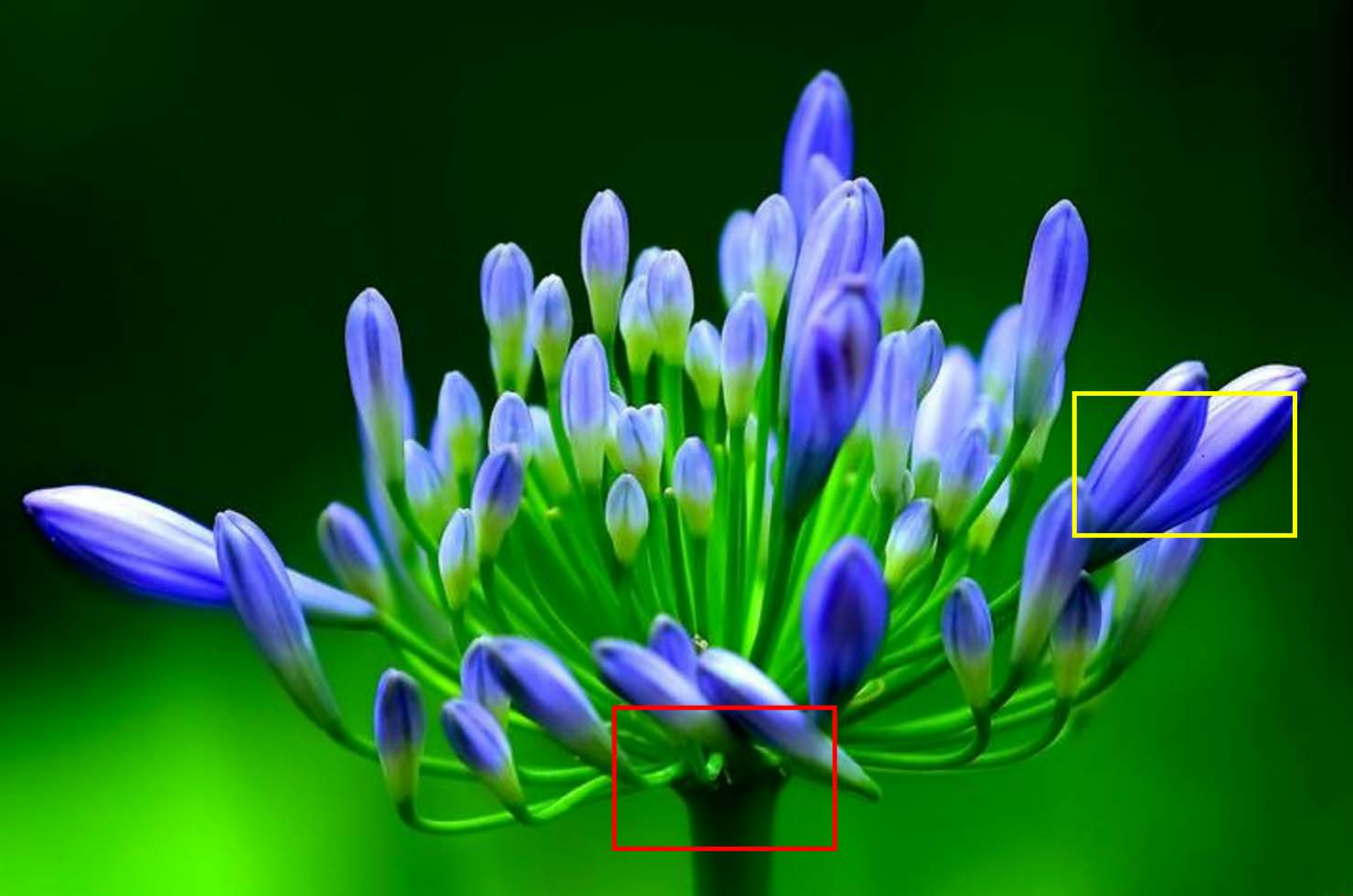}
			\begin{minipage}[b]{0.47\linewidth}
				\includegraphics[width=1\linewidth]{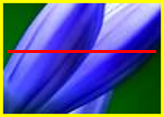}
			\end{minipage}
			\begin{minipage}[b]{0.47\linewidth}
				\includegraphics[width=1\linewidth]{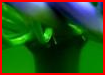}
			\end{minipage}
	\end{minipage}}	
	\subfloat{
		\begin{minipage}[b]{0.23\linewidth}
			\includegraphics[width=1\linewidth]{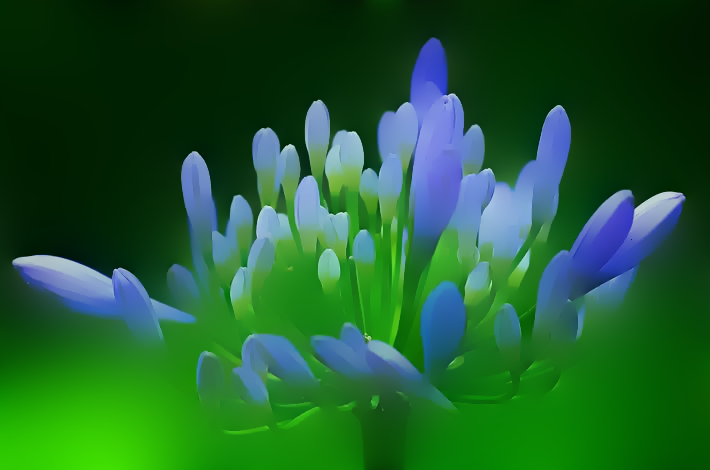}
			\begin{minipage}[b]{0.47\linewidth}
				\includegraphics[width=1\linewidth]{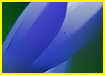}
			\end{minipage}
			\begin{minipage}[b]{0.47\linewidth}
				\includegraphics[width=1\linewidth]{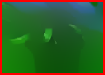}
			\end{minipage}
	\end{minipage}}
	\subfloat{
		\begin{minipage}[b]{0.23\linewidth}
			\includegraphics[width=1\linewidth]{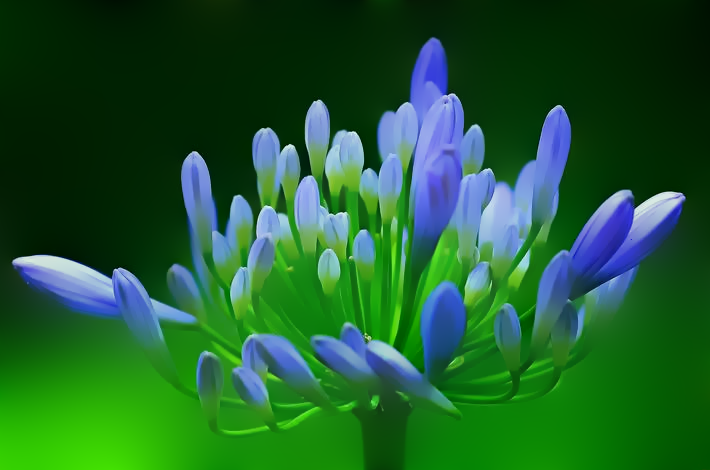}
			\begin{minipage}[b]{0.47\linewidth}
				\includegraphics[width=1\linewidth]{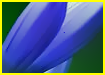}
			\end{minipage}
			\begin{minipage}[b]{0.47\linewidth}
				\includegraphics[width=1\linewidth]{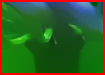}
			\end{minipage}
	\end{minipage}}
	\subfloat{
		\begin{minipage}[b]{0.23\linewidth}
			\includegraphics[width=1\linewidth]{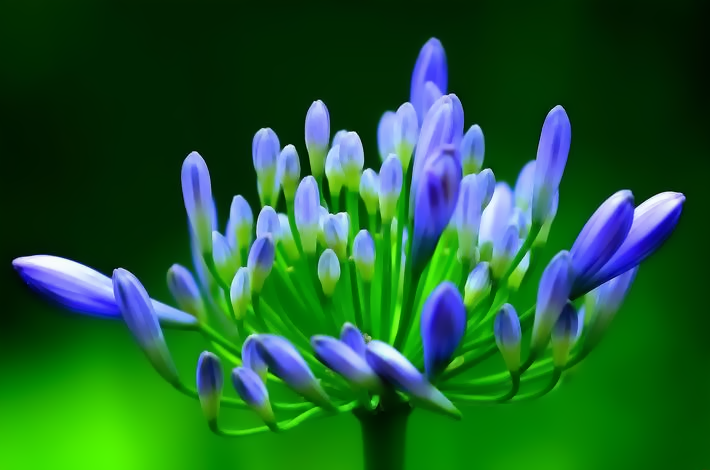}
			\begin{minipage}[b]{0.47\linewidth}
				\includegraphics[width=1\linewidth]{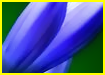}
			\end{minipage}
			\begin{minipage}[b]{0.47\linewidth}
				\includegraphics[width=1\linewidth]{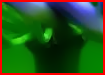}
			\end{minipage}
	\end{minipage}}
	\setcounter{subfigure}{0}
	\hfil
	\subfloat[\small Input]{
		\begin{minipage}[b]{0.23\linewidth}
			\includegraphics[width=1\linewidth]{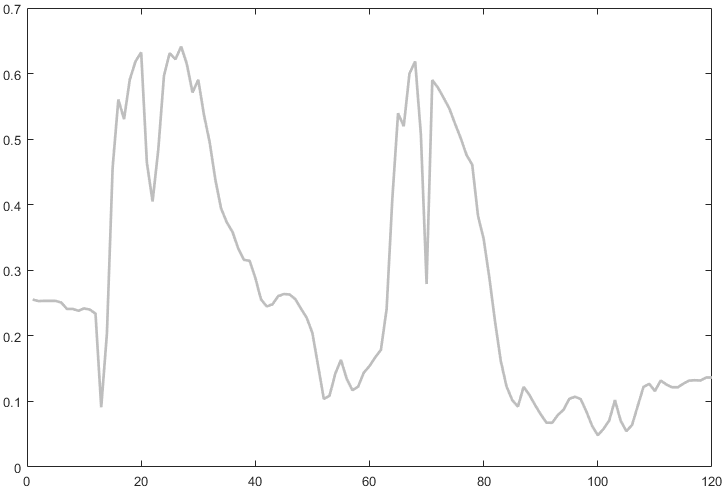}
	\end{minipage}}
	\subfloat[\small ILS(1)]{
		\begin{minipage}[b]{0.23\linewidth}
			\includegraphics[width=1\linewidth]{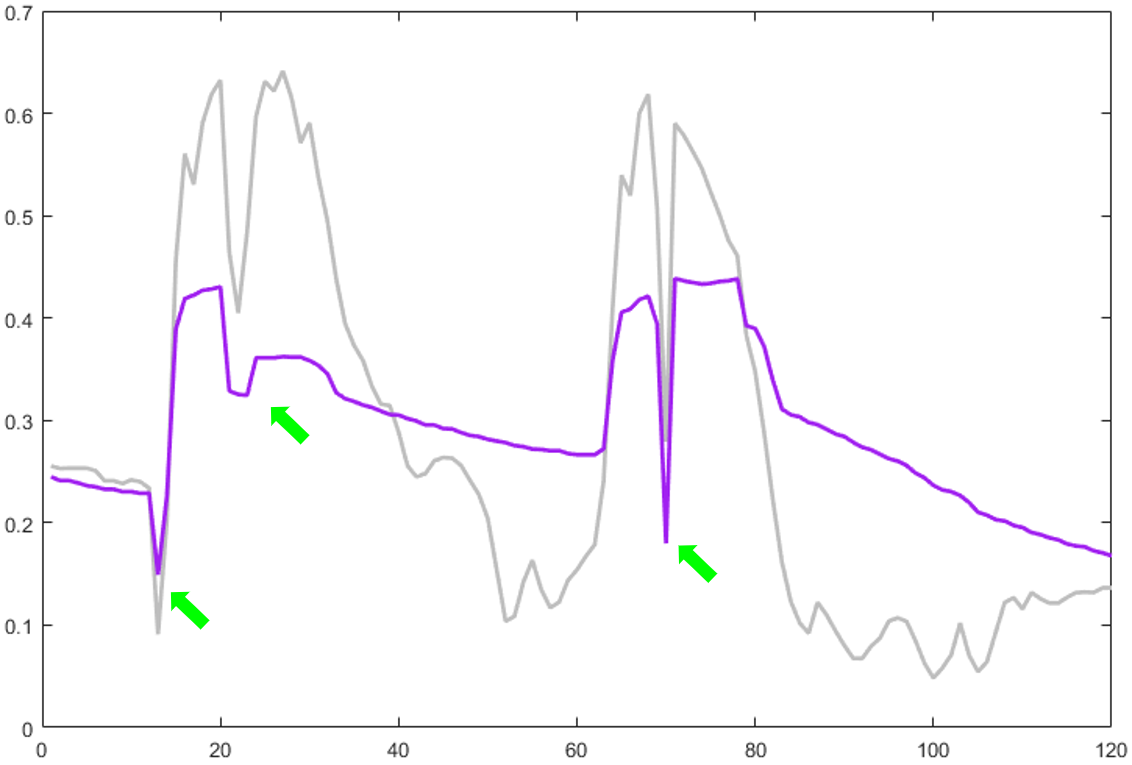}
	\end{minipage}}
	\subfloat[\small ILS(2)]{
		\begin{minipage}[b]{0.23\linewidth}
			\includegraphics[width=1\linewidth]{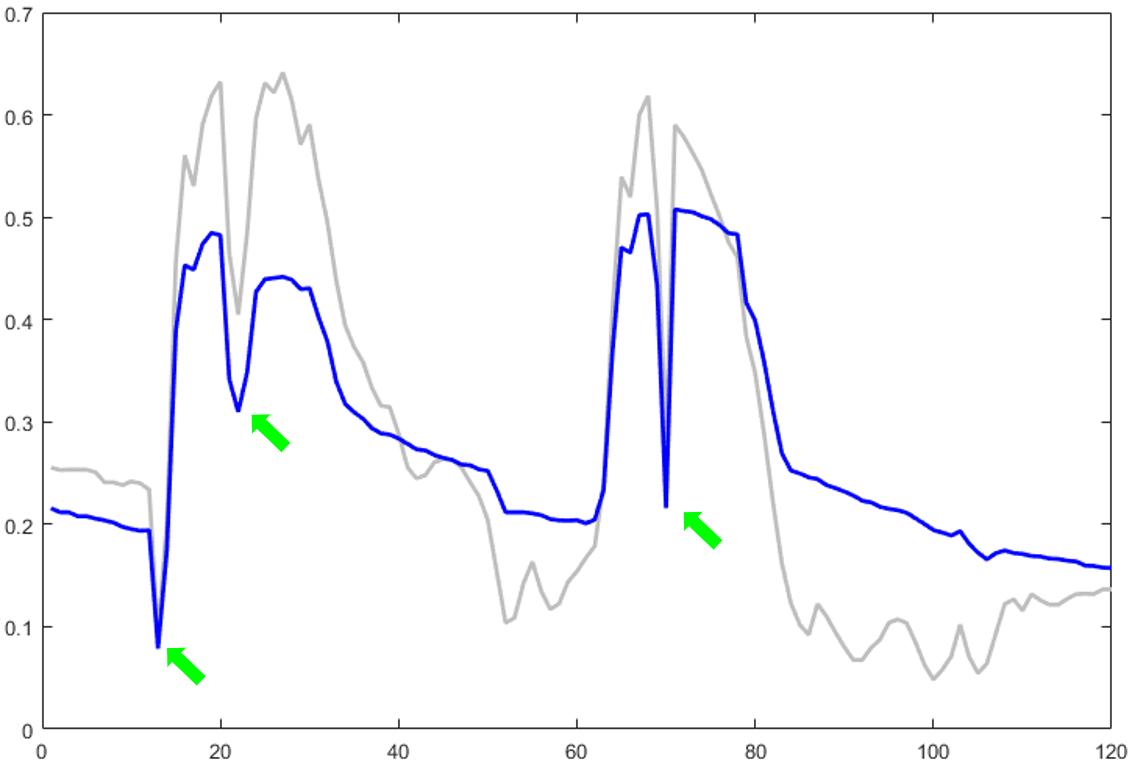}
	\end{minipage}}
	\subfloat[\small Ours]{
		\begin{minipage}[b]{0.23\linewidth}
			\includegraphics[width=1\linewidth]{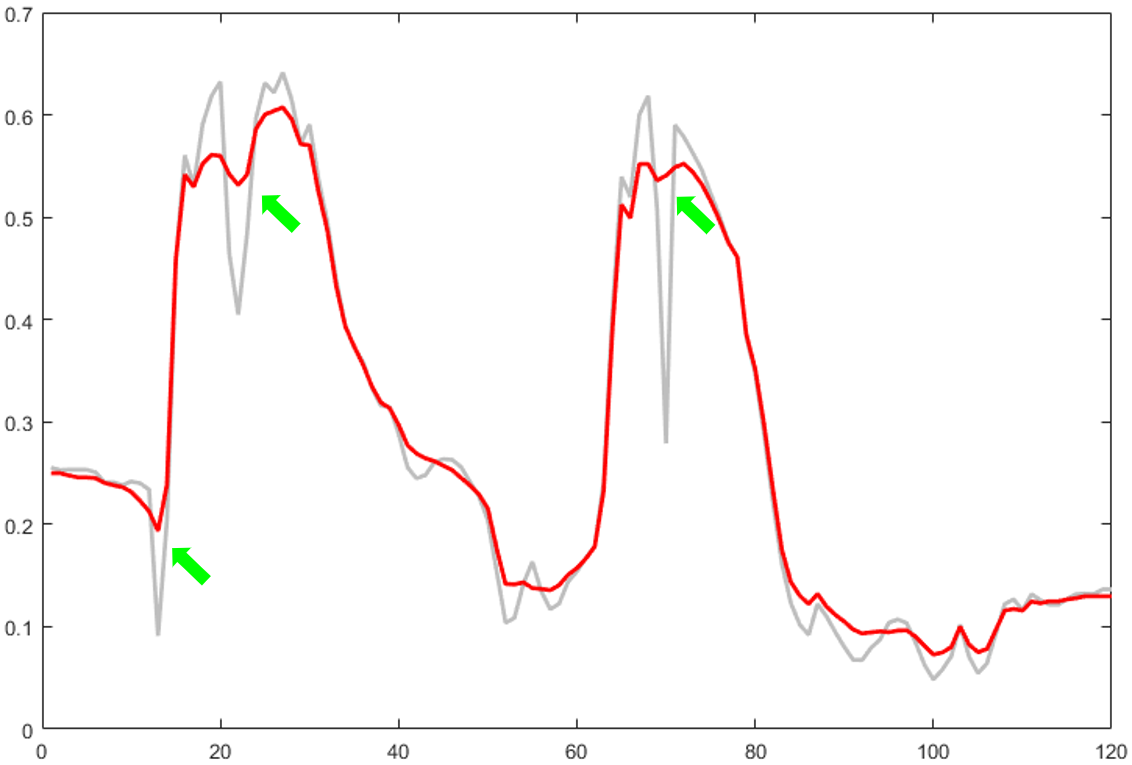}
	\end{minipage}}
	\caption{ Comparison of smoothing results by ILS and ours. (a) input image, (b) ILS ($\lambda$ = 1.0, $N$ = 10), (c) ILS ($\lambda = 3.5$, $N$ = 4), (d) Ours ($\lambda$ = 0.1, $N$ = 2, $\sigma$ = 3). Even with larger values for $\lambda$ and $N$, ILS cannot well remove high-contrast noises and this blurs some weak structures, as shown in the enlarged boxes in (b) and (c). As for ours, we can use smaller values of $\lambda$ and $N$ to well remove high-contrast noises and preserve weak structures, and so help suppressing aritifacts. This is clearly illustrated by the 1D intensity plots for a scan line in red, at the bottom of the figure. Here, the high-contrast details are marked by green arrows. (Zoom in for a better view.)}
	\label{fig6}
\end{figure}

In \cite{liu2020real}, it is discussed that larger values for $\lambda$ or $N$ would lead to stronger smoothing on the high-contrast details, but this would produce the intensity shift effect, blur weak structures, and cause compartmentalization artifacts and halo artifacts. As we can use smaller values for $\lambda$ or $N$ to smooth out high-contrast details, this is helpful for suppressing these artifacts, as illustrated in Figure \ref{fig6}. 

\subsection{Quality}
\label{subsec:quality}
\begin{figure*}[htb]
	\centering
	\subfloat[\small Input]{
		\begin{minipage}[b]{0.155\linewidth}
			\includegraphics[width=1\linewidth]{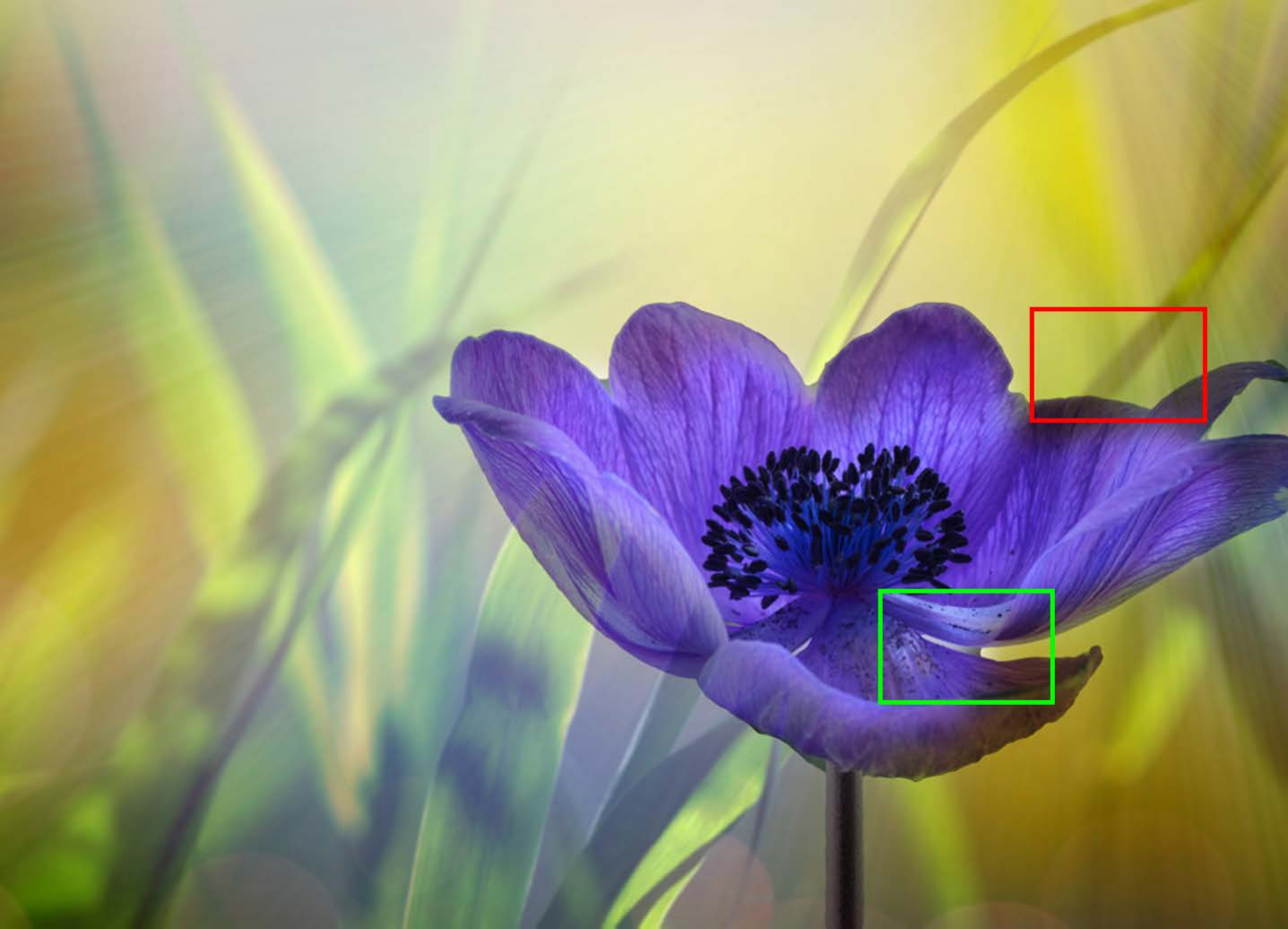}
			\begin{minipage}[b]{0.48\linewidth}
				\includegraphics[width=1\linewidth]{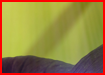}
			\end{minipage}
			\begin{minipage}[b]{0.48\linewidth}
				\includegraphics[width=1\linewidth]{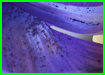}
			\end{minipage}
		\end{minipage}
		\label{fig7a}}
	\subfloat[\small BTF]{
		\begin{minipage}[b]{0.155\linewidth}
			\includegraphics[width=1\linewidth]{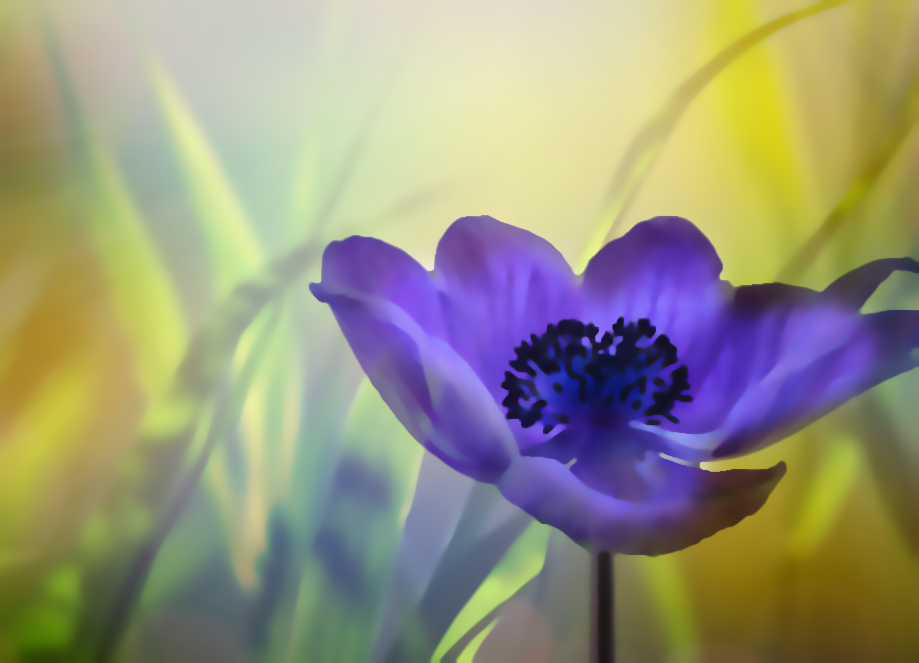}
			\begin{minipage}[b]{0.48\linewidth}
				\includegraphics[width=1\linewidth]{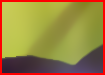}
			\end{minipage}
			\begin{minipage}[b]{0.48\linewidth}
				\includegraphics[width=1\linewidth]{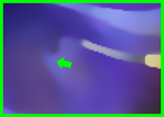}
			\end{minipage}
		\end{minipage}
		\label{fig7b}}
	\subfloat[\small IG]{
		\begin{minipage}[b]{0.155\linewidth}
			\includegraphics[width=1\linewidth]{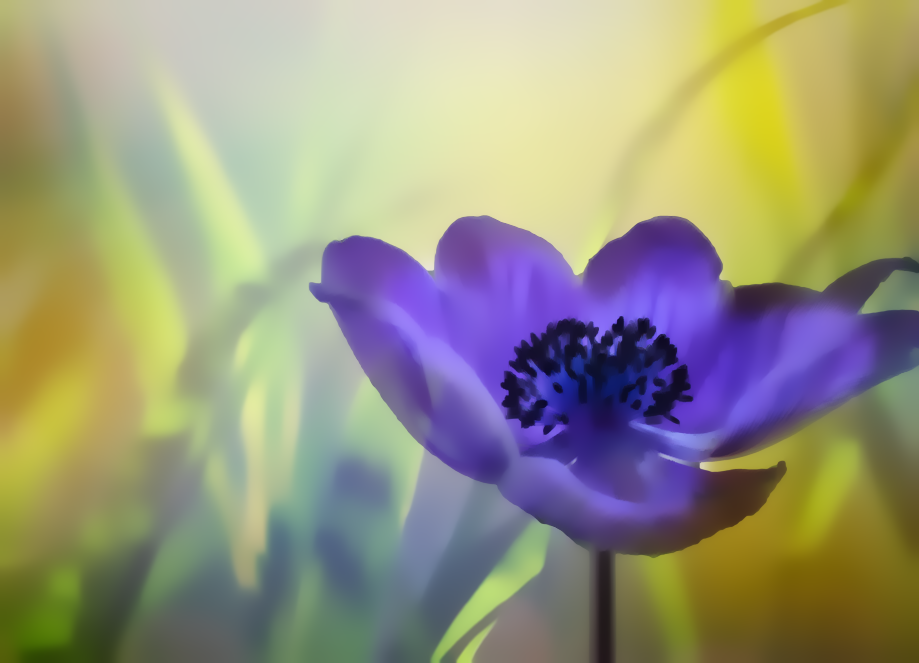}
			\begin{minipage}[b]{0.48\linewidth}
				\includegraphics[width=1\linewidth]{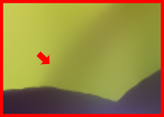}
			\end{minipage}
			\begin{minipage}[b]{0.48\linewidth}
				\includegraphics[width=1\linewidth]{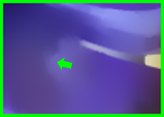}
			\end{minipage}
		\end{minipage}
		\label{fig7c}}
	\subfloat[\small EGF]{
		\begin{minipage}[b]{0.155\linewidth}
			\includegraphics[width=1\linewidth]{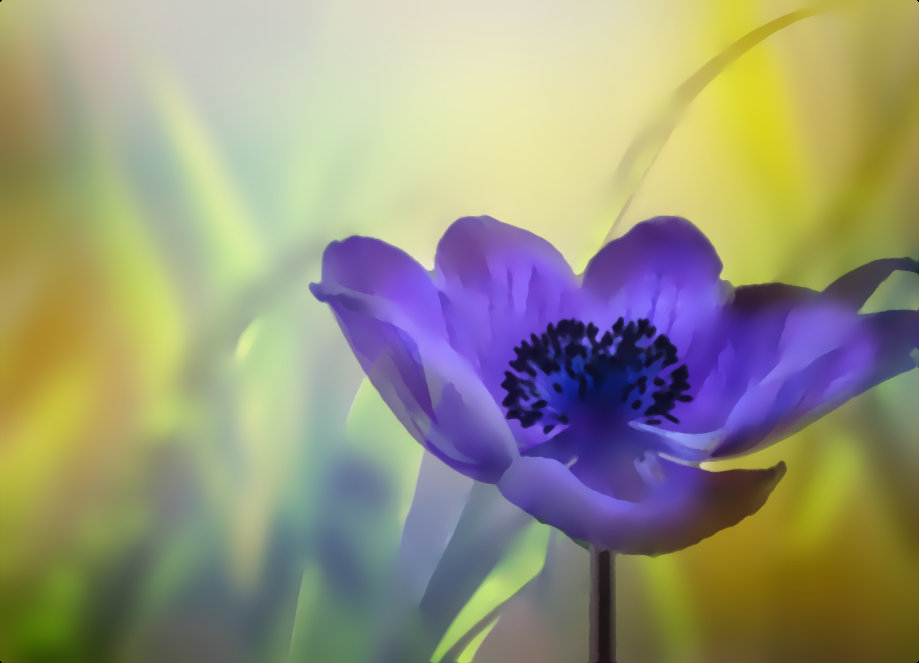}
			\begin{minipage}[b]{0.48\linewidth}
				\includegraphics[width=1\linewidth]{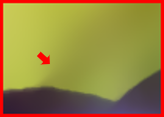}
			\end{minipage}
			\begin{minipage}[b]{0.48\linewidth}
				\includegraphics[width=1\linewidth]{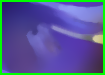}
			\end{minipage}
		\end{minipage}
		\label{fig7d}}
	\subfloat[\small EAP]{
		\begin{minipage}[b]{0.155\linewidth}
			\includegraphics[width=1\linewidth]{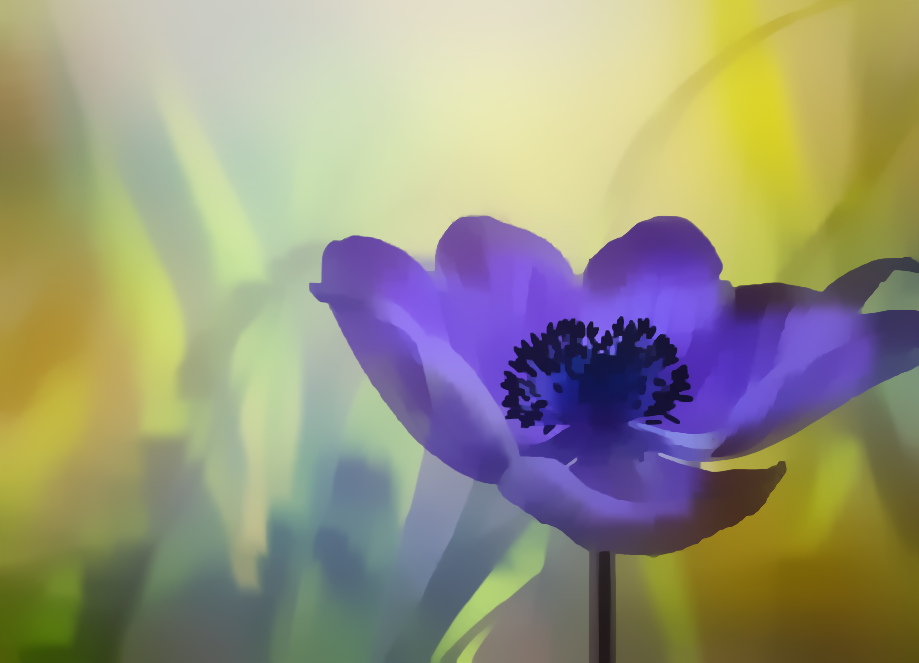}
			\begin{minipage}[b]{0.48\linewidth}
				\includegraphics[width=1\linewidth]{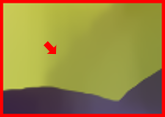}
			\end{minipage}
			\begin{minipage}[b]{0.48\linewidth}
				\includegraphics[width=1\linewidth]{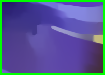}
			\end{minipage}
		\end{minipage}
		\label{fig7e}}
	\subfloat[\small ILS]{
		\begin{minipage}[b]{0.155\linewidth}
			\includegraphics[width=1\linewidth]{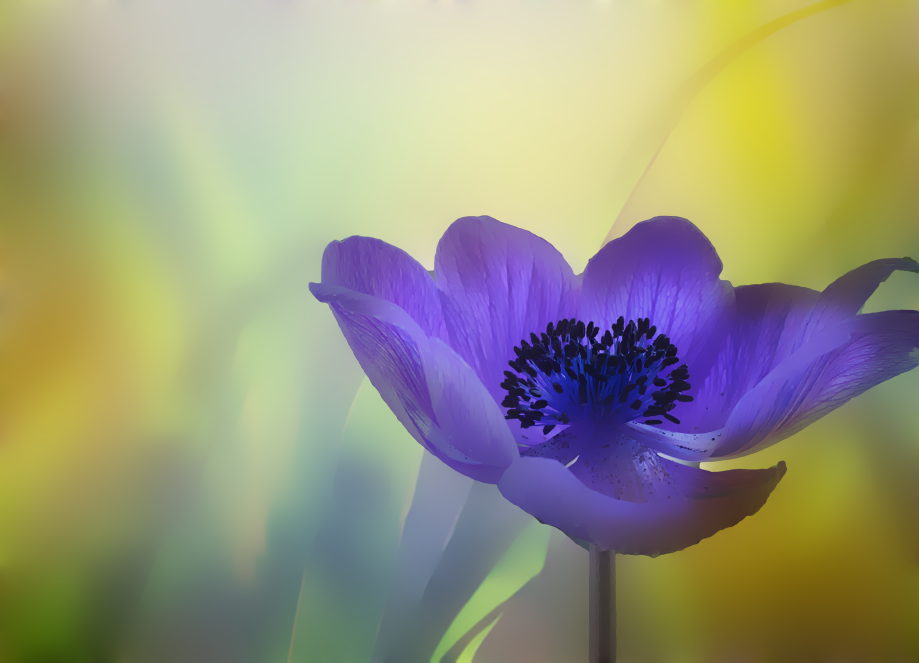}
			\begin{minipage}[b]{0.48\linewidth}
				\includegraphics[width=1\linewidth]{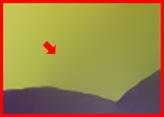}
			\end{minipage}
			\begin{minipage}[b]{0.48\linewidth}
				\includegraphics[width=1\linewidth]{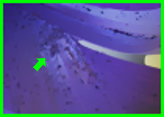}
			\end{minipage}
		\end{minipage}
		\label{fig7f}}
	\hfil
	\subfloat[\small GFES]{
		\begin{minipage}[b]{0.155\linewidth}
			\includegraphics[width=1\linewidth]{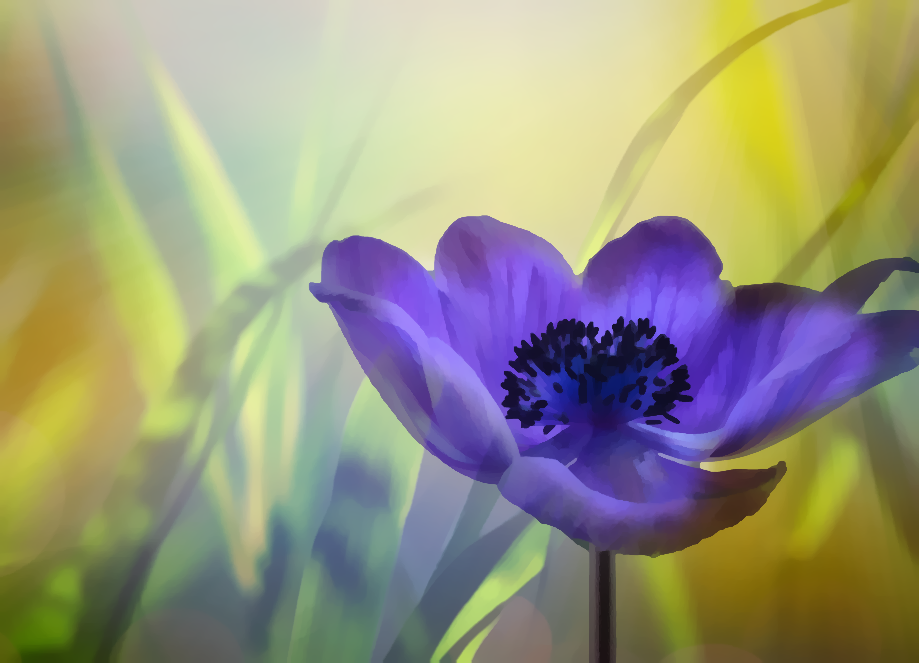}
			\begin{minipage}[b]{0.48\linewidth}
				\includegraphics[width=1\linewidth]{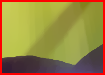}
			\end{minipage}
			\begin{minipage}[b]{0.48\linewidth}
				\includegraphics[width=1\linewidth]{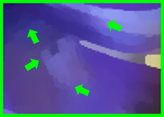}
			\end{minipage}
		\end{minipage}
		\label{fig7g}}
	\subfloat[\small STDN]{
		\begin{minipage}[b]{0.155\linewidth}
			\includegraphics[width=1\linewidth]{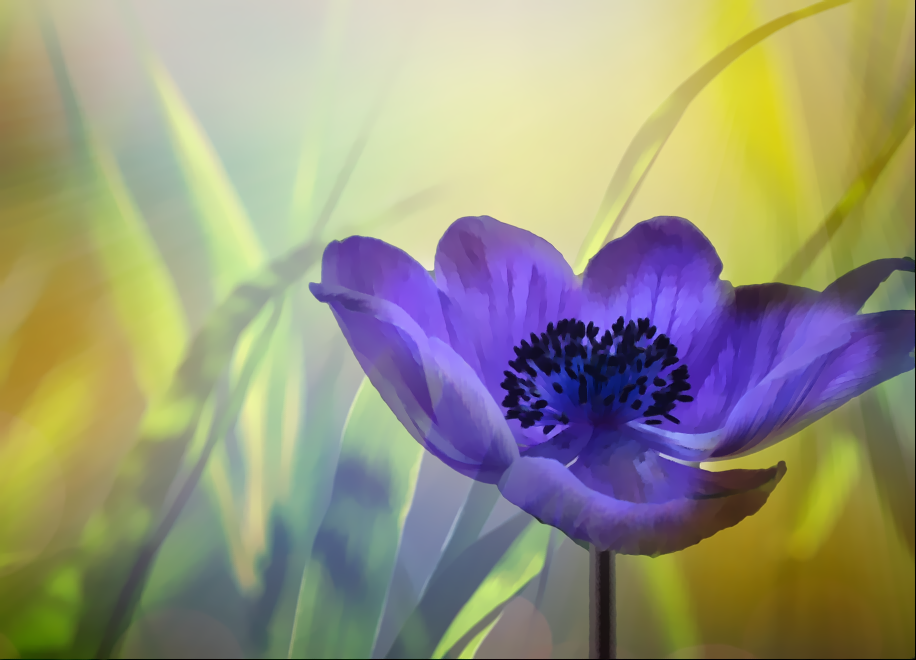}
			\begin{minipage}[b]{0.48\linewidth}
				\includegraphics[width=1\linewidth]{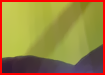}
			\end{minipage}
			\begin{minipage}[b]{0.48\linewidth}
				\includegraphics[width=1\linewidth]{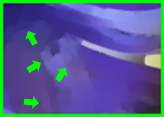}
			\end{minipage}
		\end{minipage}
		\label{fig7h}}
	\subfloat[\small Easy2Hard]{
		\begin{minipage}[b]{0.155\linewidth}
			\includegraphics[width=1\linewidth]{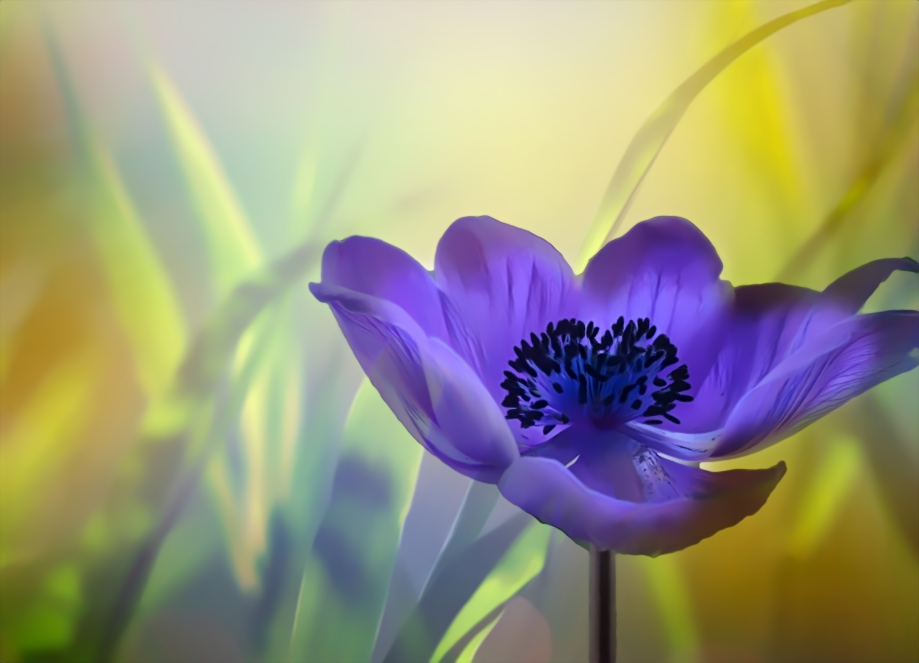}
			\begin{minipage}[b]{0.48\linewidth}
				\includegraphics[width=1\linewidth]{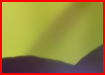}
			\end{minipage}
			\begin{minipage}[b]{0.48\linewidth}
				\includegraphics[width=1\linewidth]{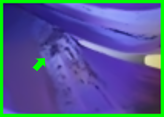}
			\end{minipage}
		\end{minipage}
		\label{fig7i}}
	\subfloat[\small DeepFSPIS]{
		\begin{minipage}[b]{0.155\linewidth}
			\includegraphics[width=1\linewidth]{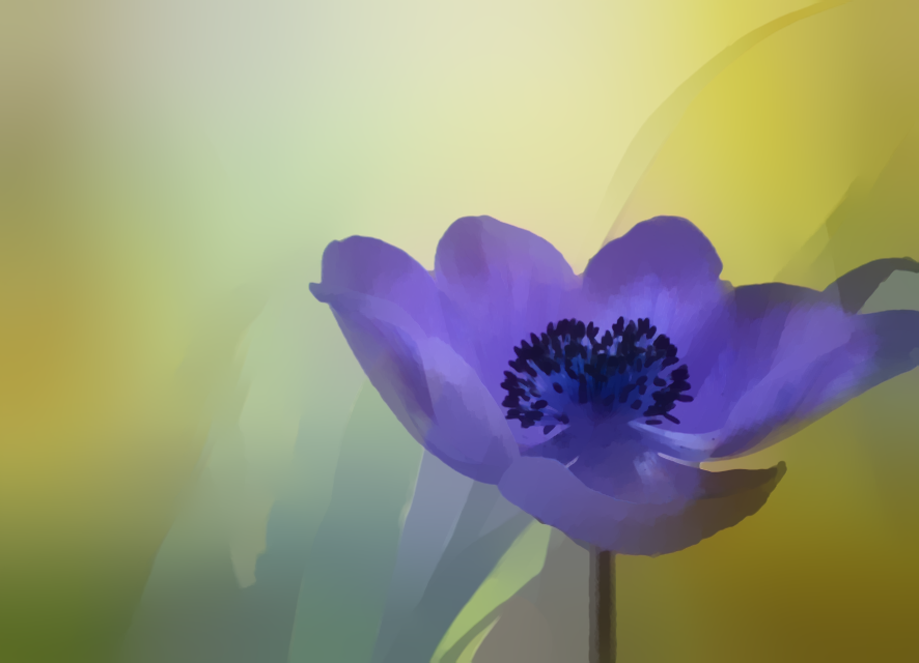}
			\begin{minipage}[b]{0.48\linewidth}
				\includegraphics[width=1\linewidth]{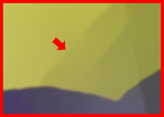}
			\end{minipage}
			\begin{minipage}[b]{0.48\linewidth}
				\includegraphics[width=1\linewidth]{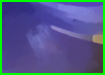}
			\end{minipage}
		\end{minipage}
		\label{fig7j}}
	\subfloat[\small CSGIS-Net]{
		\begin{minipage}[b]{0.155\linewidth}
			\includegraphics[width=1\linewidth]{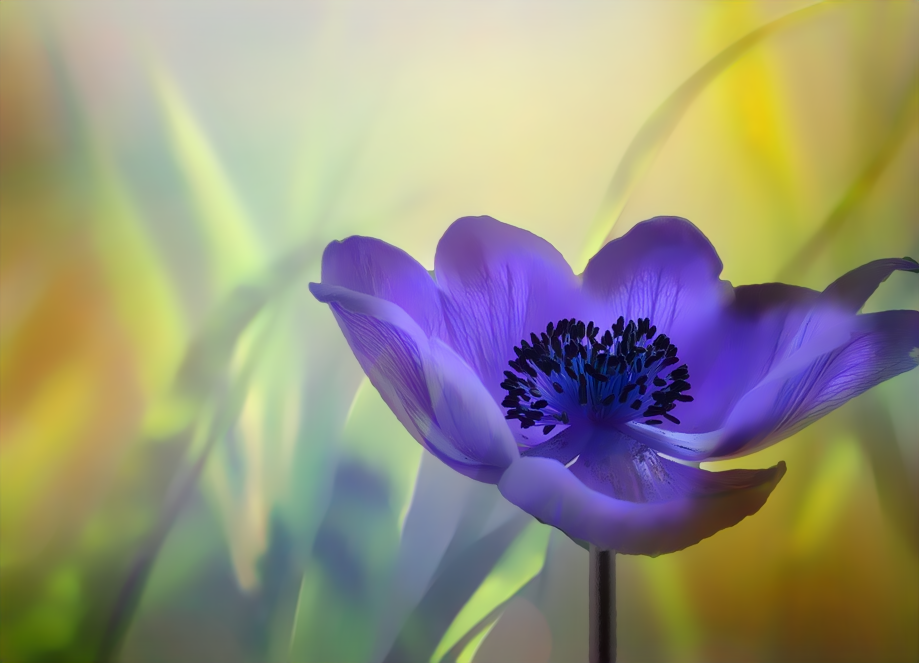}
			\begin{minipage}[b]{0.48\linewidth}
				\includegraphics[width=1\linewidth]{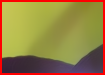}
			\end{minipage}
			\begin{minipage}[b]{0.48\linewidth}
				\includegraphics[width=1\linewidth]{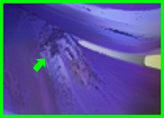}
			\end{minipage}
		\end{minipage}
		\label{fig7k}}
	\subfloat[\small Ours]{
		\begin{minipage}[b]{0.155\linewidth}
			\includegraphics[width=1\linewidth]{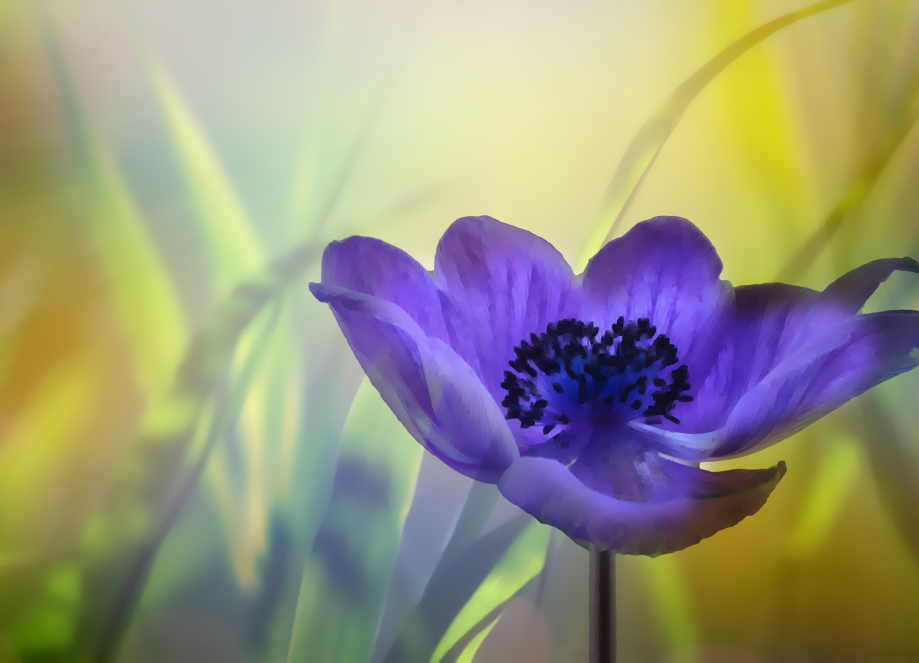}
			\begin{minipage}[b]{0.48\linewidth}
				\includegraphics[width=1\linewidth]{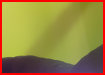}
			\end{minipage}
			\begin{minipage}[b]{0.48\linewidth}
				\includegraphics[width=1\linewidth]{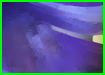}
			\end{minipage}
		\end{minipage}
		\label{fig7l}}
	\caption{Comparison of smoothing results. (a) Input image, (b) BTF \cite{cho2014bilateral} ($k$ = 5, $N$ = 4), (c) IG \cite{lee2017structure} ($\sigma$ = 4), (d) EGF \cite{sun2022edge} ($d_{0}$ = 0.1, $N$ = 1), (e) EAP \cite{zhang2020erasing} (its improved RTV method is used here, $\lambda$ = 0.015, $\sigma$ = 2, $N$ = 4), (f) ILS \cite{liu2020real} ($\lambda$ = 1.0, $N$ = 4), (g) GFES~\cite{liu2021generalized} ($\lambda$ = 0.5, $N$ = 10), (h) STDN \cite{zhou2020structure}, (i) Easy2Hard \cite{feng2021easy2hard}, (j) DeepFSPIS~\cite{li2022deep} ($\lambda$ = 0.8), (k) CSGIS-Net \cite{wang2022contrastive}, (l) Ours ($\lambda$ = 0.1, $N$ = 2, $\sigma$ = 4). We have the corresponding discussions in Subsection \ref{subsec:quality}. (Zoom in for a better view.)}
	\label{fig7}
\end{figure*}

Figure~\ref{fig7} shows the results by the methods in comparison. As can be seen, we can obtain better results than the others, especially on smoothing out high-contrast details while preserving weak structures, which are particularly shown in the enlarged boxes. More results are provided in the supplementary materials\footnote[1]{https://www.aliyundrive.com/s/rmrAZW7JQF5}.

For the filtering-based methods, BTF, IG and EGF over-smoothing structures and have the problem of preserving small weak structures, such as for the blue petals in Figure~\ref{fig7}\subref{fig7b}, \subref{fig7c} and~\subref{fig7d}, especially in the green boxes. This is due to their limited potentials for structure detection.

For the optimization-based methods, EAP and ILS have the drawback of blurring small structures, as shown in red boxes in Figure~\ref{fig7}\subref{fig7e} and~\subref{fig7f}. Besides, ILS cannot remove high-contrast details, as shown in the green box in Figure \ref{fig7}\subref{fig7f}. GFES and STDN can obtain better smoothing results, but they produce patch-like appearances, as shown in the green boxes in  Figure~\ref{fig7}\subref{fig7g} and~\subref{fig7h}. This is because they cannot well distinguish neighboring structures and details with similar intensities due to their using truncated Huber penalty function or learning techniques, and so have them handled similarly to cause block effects.

For the learning-based methods, Easy2Hard and CSGIS-Net fail to remove high-contrast details, as shown in the green boxes in Figure \ref{fig7}\subref{fig7i} and~\subref{fig7k}. DeepFSPIS cannot preserve weak structures, as shown in the red box in Figure~\ref{fig7}\subref{fig7j}. They are prevented by their training data, as discussed in Section \ref{sec:relateWork}.

For quantitative evaluation of image smoothing results, it still lacks effective measures, as discussed in~\cite{liu2020real}. We will indirectly show our improvements by quantitative evaluation of tone mapping results based on the smoothed results with GFES, ILS and our method, to be discussed in Subsection~\ref{subsec:app}. More comparisons with previous methods have been discussed in~\cite{liu2020real,liu2021generalized} to show the superiority of GFES and ILS over previous methods.

\subsection{Efficiency}
\label{subsec:efficien}
In comparison with ILS, our improved method only differs from it on the computation of the penalty function. Here, we add some weight computation. With a simple investigation, it is known that we don’t increase much time cost in each iteration. By the statistics on time cost in Table~\ref{tab:table1}, it is known ours is much faster than ILS, no matter whether on CPUs or on GPUs. This is much benefited from our reduced iterations. As for other compared methods, ours is much faster than theirs. The statistics in Table~\ref{tab:table1} show that we can perform real time image smoothing on GPUs in handling ordinary images. This is much superior to existing methods.

\begin{table}[htb]
	\caption{Running time (in seconds) for the tested methods to produce the results in Figure \ref{fig1} (615 * 461 pixels), Figure \ref{fig7} (919*663 pixels) and Figure  \ref{fig8} (800 * 533 pixels).}
	\label{tab:table1}
	\centering
	\scalebox{0.75}{
		\begin{tabular}{lllllll}
			\toprule
			\multicolumn{1}{c}{\multirow{2}{*}{Methods}} & \multicolumn{2}{c}{Figure\ref{fig1}\subref{fig1a}}      & \multicolumn{2}{c}{Figure\ref{fig7}\subref{fig7a}}      & \multicolumn{2}{c}{Figure\ref{fig8}\subref{fig8a}}             \\ \cline{2-7} 
			\multicolumn{1}{c}{}       & \multicolumn{1}{c}{CPU} & \multicolumn{1}{c}{GPU} & \multicolumn{1}{c}{CPU} & \multicolumn{1}{c}{GPU} & \multicolumn{1}{c}{CPU} & \multicolumn{1}{c}{GPU} \\ 
			\midrule 
			BTF         & 10.459s     & /        & 18.046s       & /         & 15.035s      & /              \\
			IG          & 1.657s      & 0.125s   & 3.382s        & 0.169s    & 2.253s       & 0.146s         \\
			EGF         & 12.325s     & /        & 14.284s       & /         & 18.718s      & /               \\
			CSGIS-Net   & /           & 1.084s   & /             & 1.586s    & /            & 1.242s               \\
			Easy2Hard   & /           & 1.074s   & /             & 1.595s    & /            & 1.234s               \\
			DeepFSPIS   & /           & 0.710s   & /             & 0.963s    & /            & 0.845s               \\
			STDN        & 136.395s    & /        & 299.463s      & /         & 208.494s     & /                \\
			EAP         & 10.421s     & /        & 23.844s       & /         & 18.469s      & /               \\
			GFES        & 7.491s      & /        & 14.544s       & /         & 10.452s      & /              \\
			ILS         & 0.909s      & 0.058s   & 0.818s        & 0.063s    & 0.551s       & 0.041s          \\
			Ours        & 0.273s      & 0.026s   & 0.569s        & 0.052s    & 0.368s       & 0.032s         \\ 
			\bottomrule
	\end{tabular}}
\end{table}

\subsection{Applications }
\label{subsec:app}

\begin{figure*}[htb]
	\centering
	\subfloat[\small Input]{
		\begin{minipage}[b]{0.19\linewidth}
			\includegraphics[width=1\linewidth]{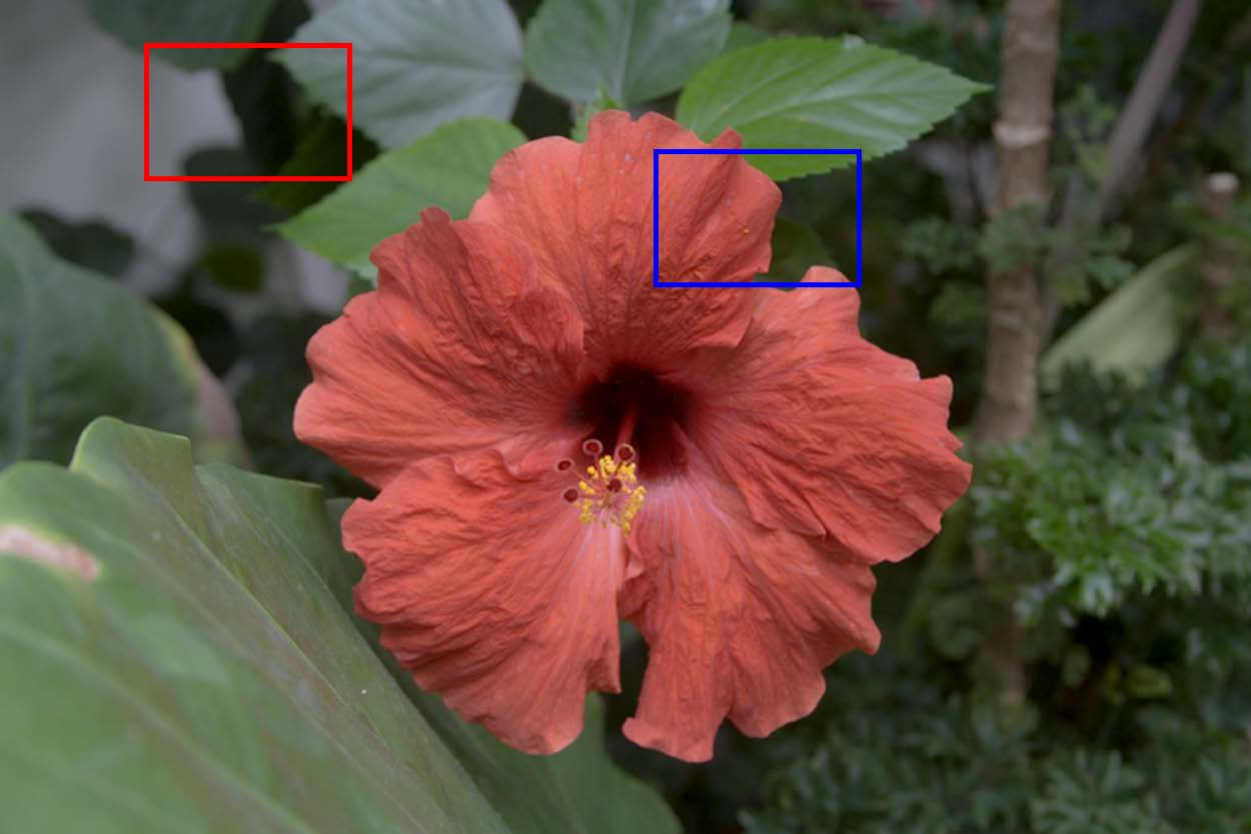}
			\begin{minipage}[b]{0.48\linewidth}
				\includegraphics[width=1\linewidth]{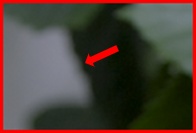}
			\end{minipage}
			\begin{minipage}[b]{0.48\linewidth}
				\includegraphics[width=1\linewidth]{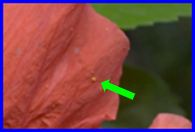}
			\end{minipage}
		\end{minipage}
		\label{fig8a}}	
	\subfloat[\small ILS smoothed]{
		\begin{minipage}[b]{0.19\linewidth}
			\includegraphics[width=1\linewidth]{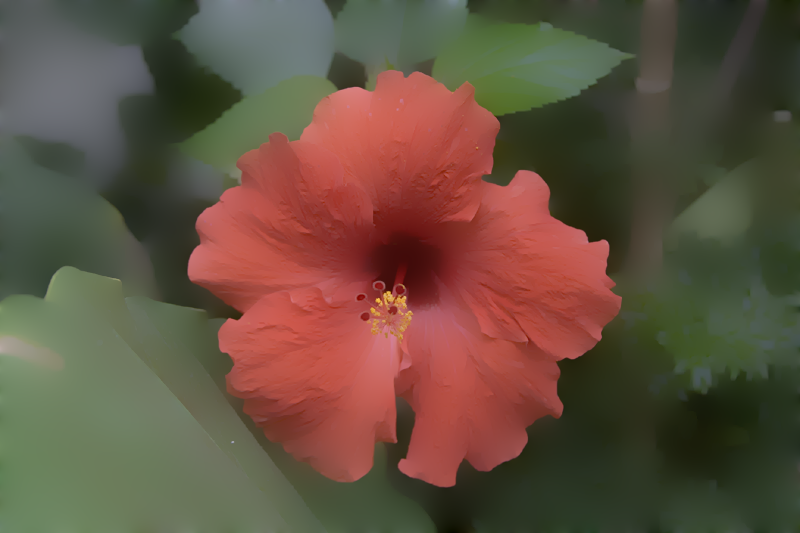}
			\begin{minipage}[b]{0.48\linewidth}
				\includegraphics[width=1\linewidth]{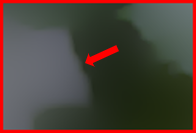}
			\end{minipage}
			\begin{minipage}[b]{0.48\linewidth}
				\includegraphics[width=1\linewidth]{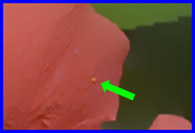}
			\end{minipage}
	\end{minipage}}
	\subfloat[\small ILS enhanced]{
		\begin{minipage}[b]{0.19\linewidth}
			\includegraphics[width=1\linewidth]{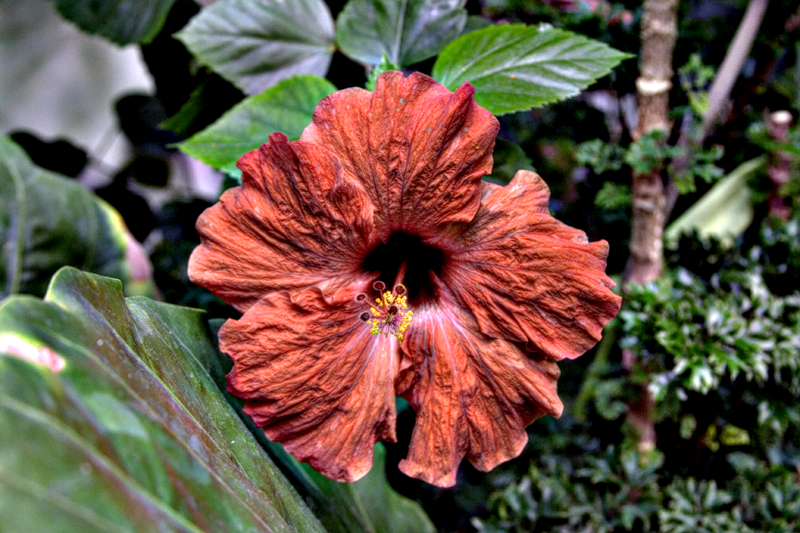}
			\begin{minipage}[b]{0.48\linewidth}
				\includegraphics[width=1\linewidth]{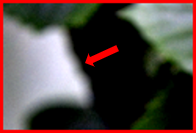}
			\end{minipage}
			\begin{minipage}[b]{0.48\linewidth}
				\includegraphics[width=1\linewidth]{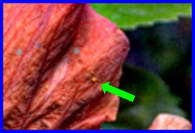}
			\end{minipage}
	\end{minipage}}
	\subfloat[\small Ours smoothed]{
		\begin{minipage}[b]{0.19\linewidth}
			\includegraphics[width=1\linewidth]{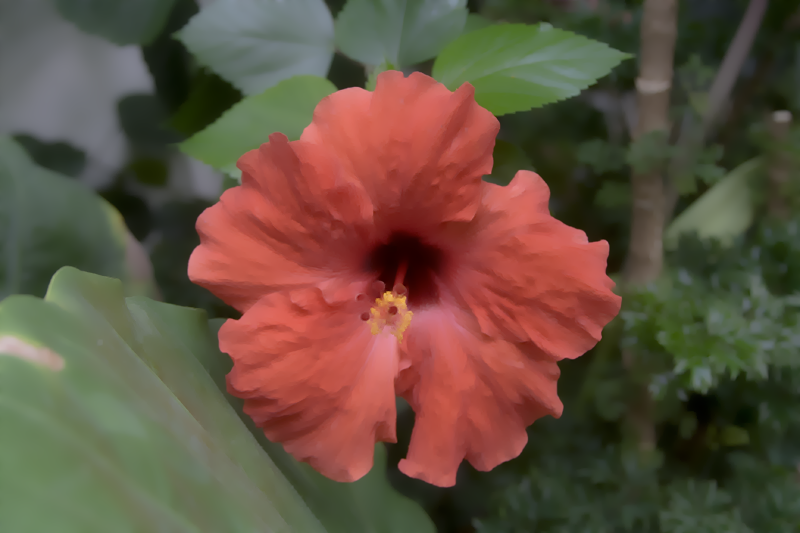}
			\begin{minipage}[b]{0.48\linewidth}
				\includegraphics[width=1\linewidth]{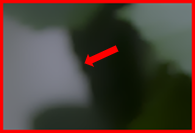}
			\end{minipage}
			\begin{minipage}[b]{0.48\linewidth}
				\includegraphics[width=1\linewidth]{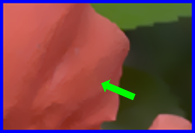}
			\end{minipage}
	\end{minipage}}
	\subfloat[\small Ours enhanced]{
		\begin{minipage}[b]{0.19\linewidth}
			\includegraphics[width=1\linewidth]{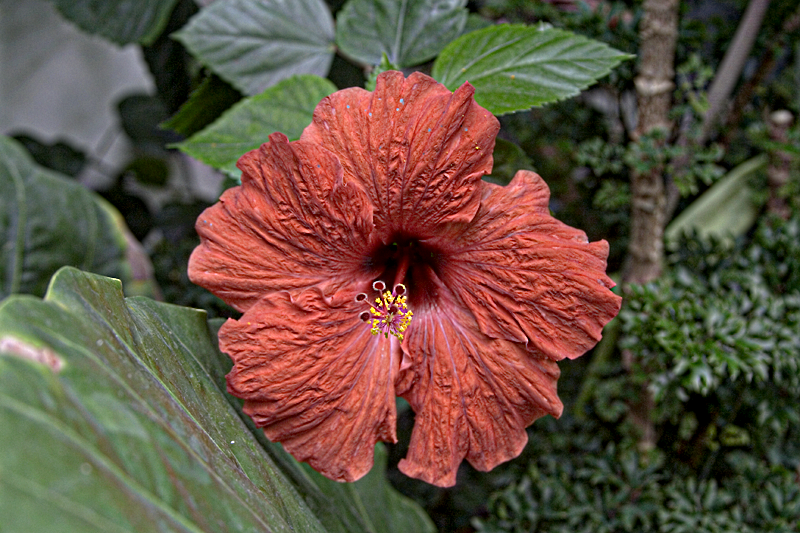}
			\begin{minipage}[b]{0.48\linewidth}
				\includegraphics[width=1\linewidth]{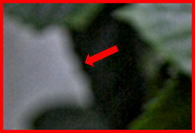}
			\end{minipage}
			\begin{minipage}[b]{0.48\linewidth}
				\includegraphics[width=1\linewidth]{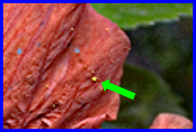}
			\end{minipage}
	\end{minipage}}
	\caption{Comparison on avoiding compartmentalization artifacts in detail enhancement. As can be seen in (b) and (d), our smoothed results are better in preserving structures and removing high-contrast details. In (c) and (e) for detail enhancement with 3x detail, ours effectively alleviates the compartmentalization artifacts, as marked by arrows in the red boxes, while achieving better results, as marked by arrows in the blue boxes. Parameters: (b) ILS ($\lambda$ = 1.0, $N$ = 4), (d) Ours ($\lambda$ = 0.1, $N$ = 2, $\sigma$ = 5). }
	\label{fig8}
\end{figure*}

\begin{figure}[htb]
	\centering
	\subfloat[\small Input]{
		\begin{minipage}[b]{0.225\linewidth}
			\includegraphics[width=1\linewidth]{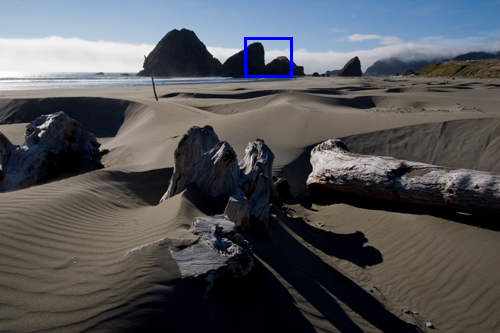}
			\begin{minipage}[b]{0.47\linewidth}
				\includegraphics[width=1\linewidth]{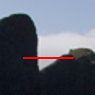}
			\end{minipage}
			\begin{minipage}[b]{0.47\linewidth}
				\includegraphics[width=1\linewidth]{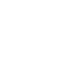}
			\end{minipage}
		\end{minipage}
		\label{fig9a}}	
	\subfloat[\small EGF]{
		\begin{minipage}[b]{0.225\linewidth}
			\includegraphics[width=1\linewidth]{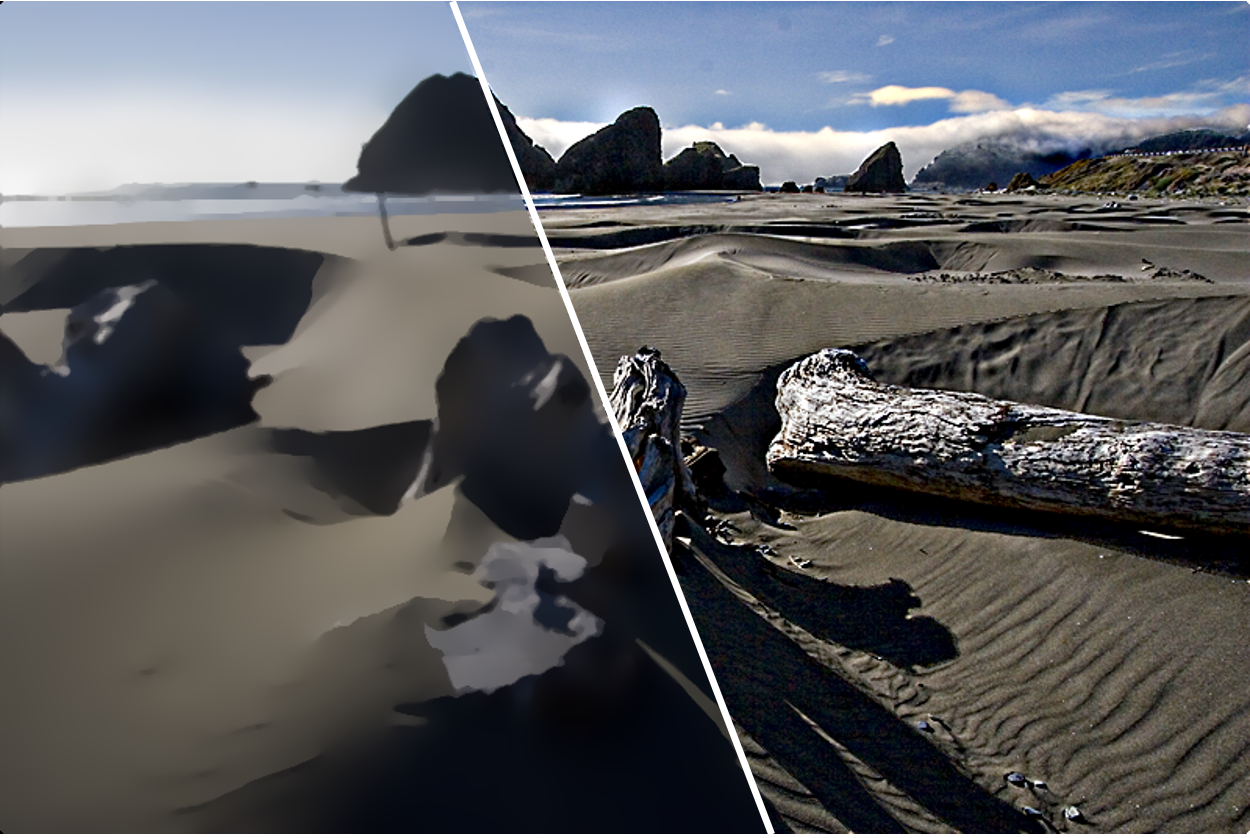}
			\begin{minipage}[b]{0.47\linewidth}
				\includegraphics[width=1\linewidth]{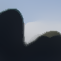}
			\end{minipage}
			\begin{minipage}[b]{0.47\linewidth}
				\includegraphics[width=1\linewidth]{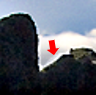}
			\end{minipage}
		\end{minipage}
		\label{fig9b}}
	\subfloat[\small ILS]{
		\begin{minipage}[b]{0.225\linewidth}
			\includegraphics[width=1\linewidth]{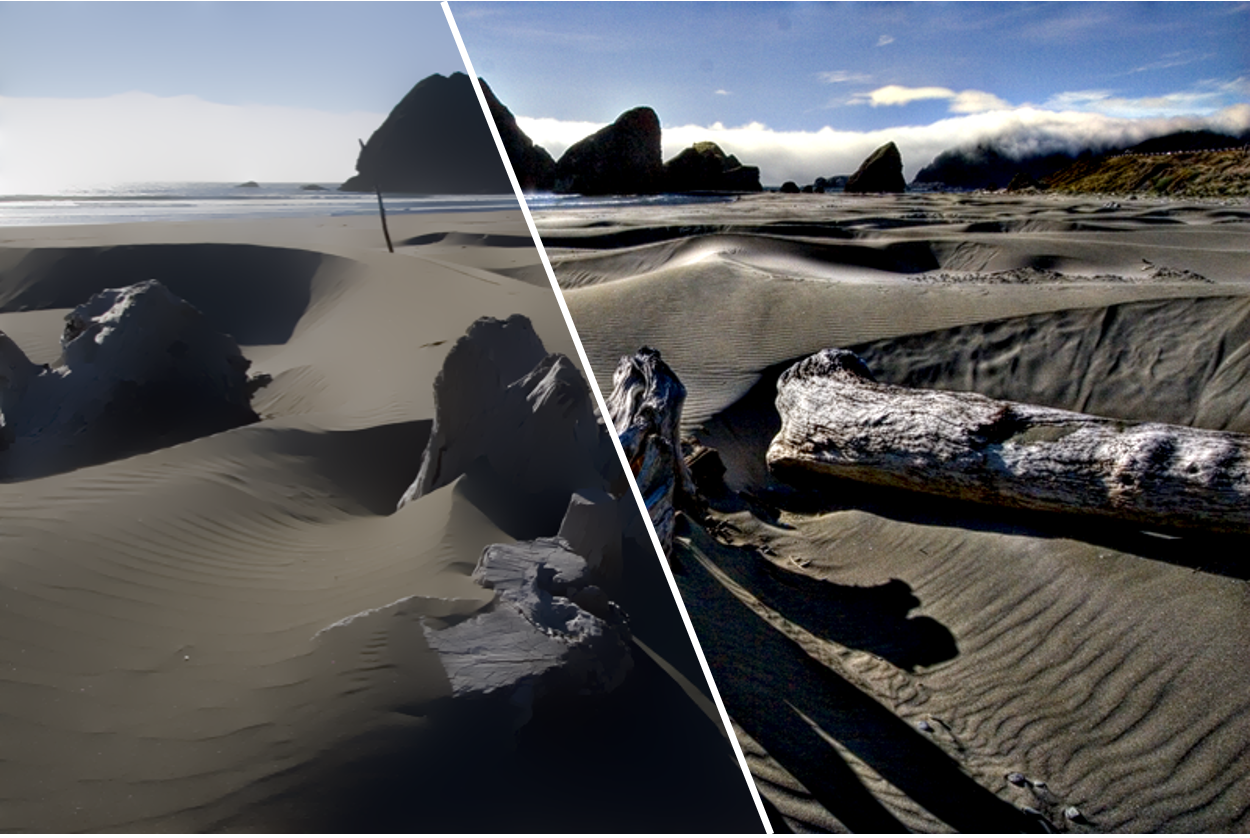}
			\begin{minipage}[b]{0.47\linewidth}
				\includegraphics[width=1\linewidth]{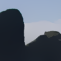}
			\end{minipage}
			\begin{minipage}[b]{0.47\linewidth}
				\includegraphics[width=1\linewidth]{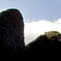}
			\end{minipage}
		\end{minipage}
		\label{fig9c}}
	\subfloat[\small Ours]{
		\begin{minipage}[b]{0.225\linewidth}
			\includegraphics[width=1\linewidth]{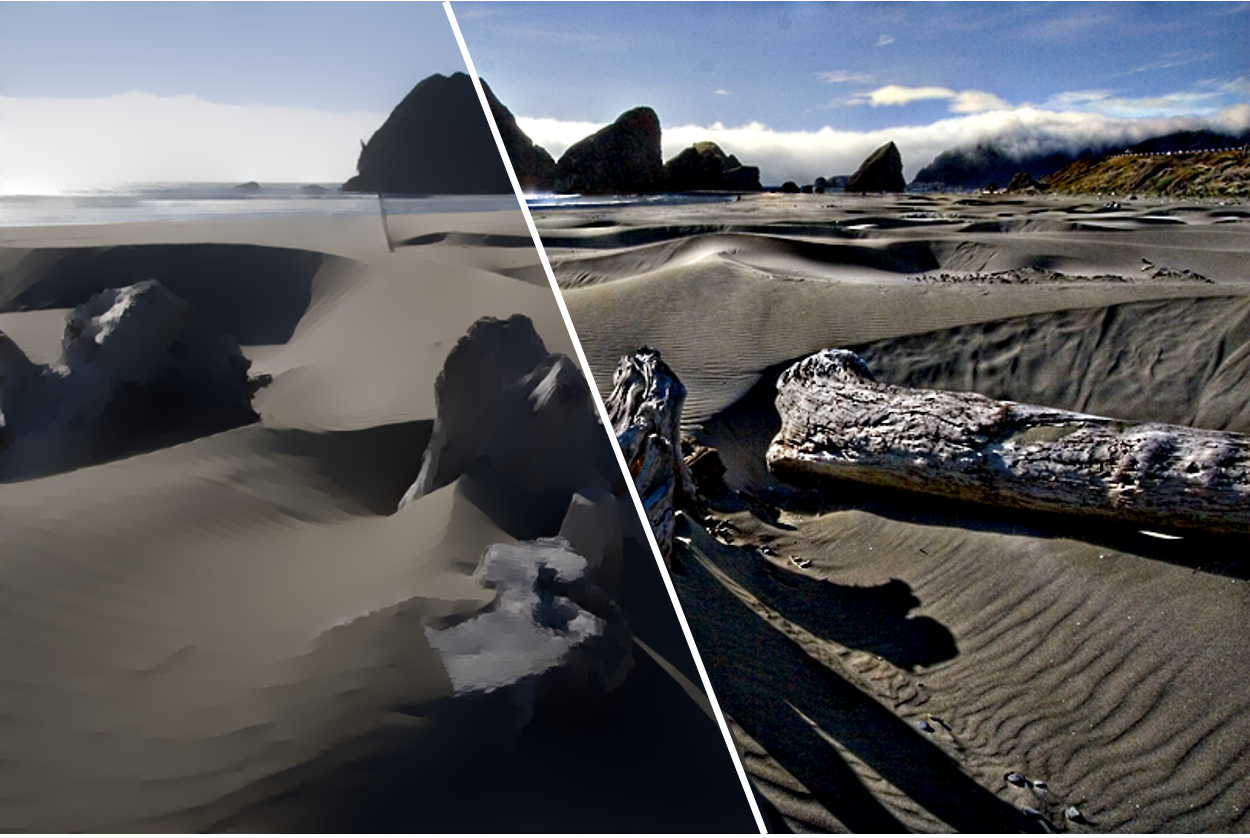}
			\begin{minipage}[b]{0.47\linewidth}
				\includegraphics[width=1\linewidth]{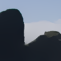}
			\end{minipage}
			\begin{minipage}[b]{0.47\linewidth}
				\includegraphics[width=1\linewidth]{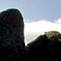}
			\end{minipage}
		\end{minipage}
		\label{fig9d}}
	\caption{ Comparison of avoiding halo artifacts in detail enhancement. (a) input image, (b) EGF \cite{sun2022edge} ($d_{0}$ = 0.1, $N$ = 2), (c) ILS \cite{liu2020real} ($\lambda$ = 1.0, $N$ = 4), (d) Ours ($\lambda$ = 1.0, $N$ = 3, $\sigma$ = 4). Top: the left parts in (b) $\sim$ (d) are the smoothed images and the right parts are the corresponding 3× detail enhanced images. Bottom: Close-up of the blue box in (a) and the corresponding smoothed results (the left) and enhanced results (the right). The halo effects in (b) are very evident, as marked by the red arrow. As for ILS, it has similar results as ours, but it has stronger halo effects than ours, as illustrated by the 1D intensity plots in Figure \ref{fig10}. }
	\label{fig9}
\end{figure}

Our method has better performance on preserving structures and removing details than existing methods. Thus, we can improve many applications, as illustrated in Figure \ref{fig8} and Figure \ref{fig9} for detail enhancement, and in Figure \ref{fig11} for HDR tone mapping. More results are provided in the supplementary materials\footnote[2]{https://www.aliyundrive.com/s/rmrAZW7JQF5}.

\begin{figure}[htb]
	\centering
	\begin{minipage}{0.6\linewidth}
		\includegraphics[width=\linewidth]{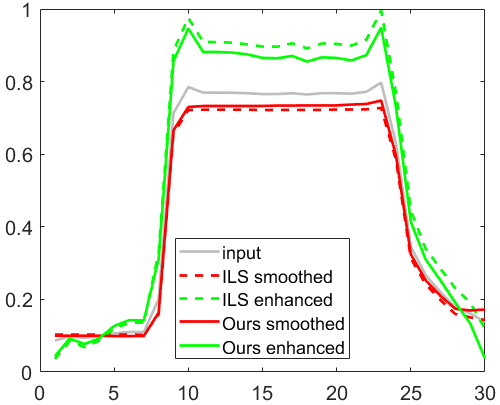}
	\end{minipage}
	\caption{ The 1D intensity plots of the red line in Figure \ref{fig9}\subref{fig9a} for the results of ILS and ours in Figure \ref{fig9}. Compared with ILS, We can have results  nearer to the original signals, and so more effective to suppress artifacts. }
	\label{fig10}
\end{figure}

For quantitative evaluation of our improvements, we made an investigation on the HDR tone mapping results based on the smoothed images produced by GFES, ILS and ours. Here, the 25 HDR images used in ~\cite{liu2020real,liu2021generalized} are used and the tone mapping quality index (TMQI) proposed by Yeganeh et al.~\cite{yeganeh2012objective} is used for quantitative evaluation. TMQI first evaluates the structural fidelity and the naturalness of the tone mapping images, and then combines these two measurements with a power function to give a final score ranging from 0.0 to 1.0. Larger values of TMQI indicate better quality, and vice versa. In Table ~\ref{rebuttal_tab1}, it is listed the statistics about the average evaluation results for the 25 images, which show our superiority over ILS and GFES.

\begin{table}[htb]
	\centering
	\caption{Quantitative evaluation for HDR tone mapping results.}
	\label{rebuttal_tab1}
	\scalebox{0.8}{
		\begin{tabular}{lcccc}
			\toprule     & GFES(EP-1) & GFES(SP-2) & ILS & Ours \\
			\midrule  Naturalness & 0.4463 & 0.4703 & 0.5003 & 0.5109 \\
			Fidelity    & 0.8461 & 0.8451 & 0.8637 & 0.8695 \\
			TMQI        & 0.8666 & 0.8712 & 0.8753 & 0.8806 \\
			\bottomrule
	\end{tabular}}
	\begin{threeparttable}[b]
		\begin{tablenotes}
			\item[]Note: EP-1 and SP-2 are two models for using GFES. Here, the data for GFES(EP-1), GFES(SP-2) and ILS are from their corresponding published papers~\cite{liu2021generalized,liu2020real}.
		\end{tablenotes}
	\end{threeparttable}
\end{table}

{\bf Limitation.} Our current implementation is by using interval gradients to distinguish details from structures, and so execute our weight computation. As the potentials of interval gradients are not very strong in distinguishing structures from the others, our results would suffer from this, as shown in Figure~\ref{fig12}. As we know, texture filtering is required in many applications. For this, image smoothing methods cannot be directly used to texture filtering as textures can sometimes have very high contrast. Thus, for image smoothing methods to smooth out texture details,  a preprocessing is often required to smooth the input image for reducing contrasts of textures, like done with ILS \cite{liu2020real}. As for our method, we can take the same way to perform texture filtering. 

\begin{figure}[htb]
	\centering
	\subfloat[\small GFES(EP-1)]{
		\begin{minipage}[b]{0.45\linewidth}
			\includegraphics[width=1\linewidth]{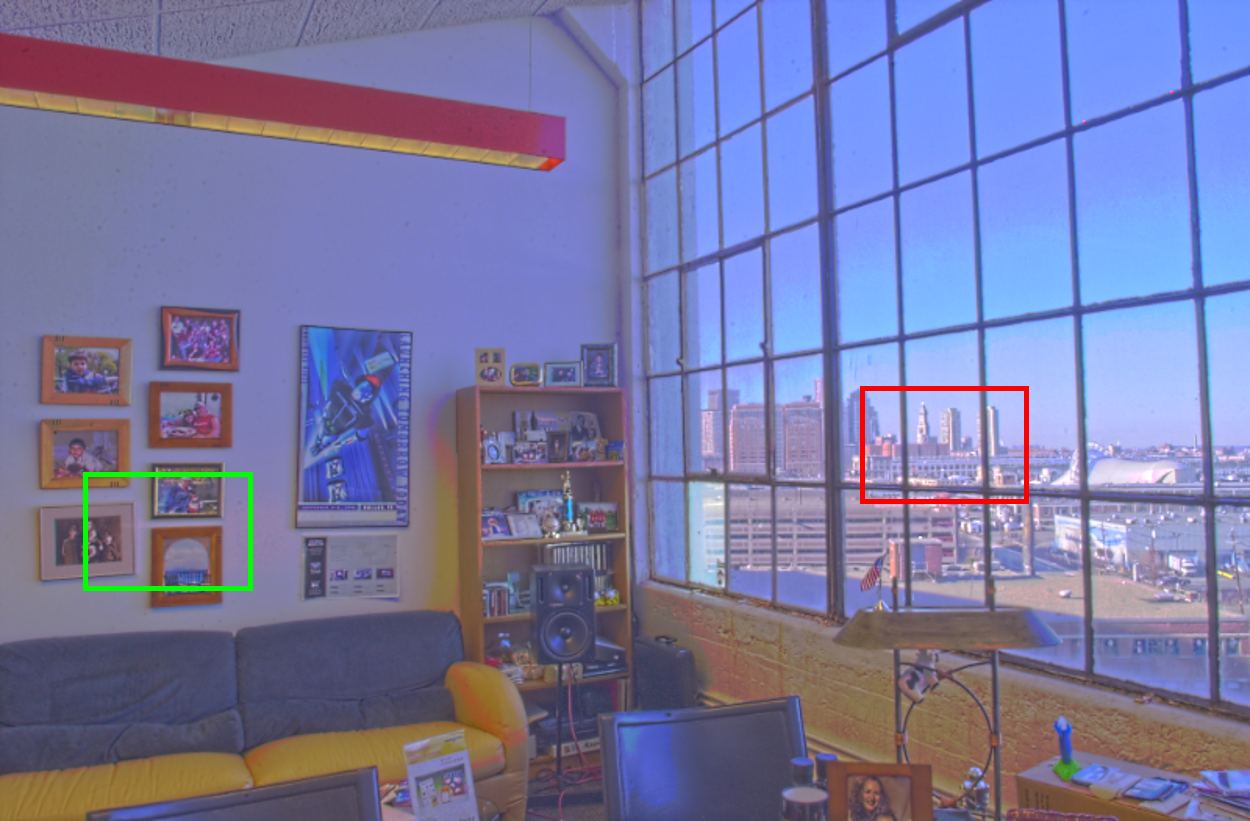}
			\begin{minipage}[b]{0.485\linewidth}
				\includegraphics[width=1\linewidth]{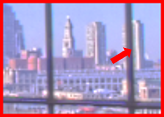}
			\end{minipage}
			\begin{minipage}[b]{0.485\linewidth}
				\includegraphics[width=1\linewidth]{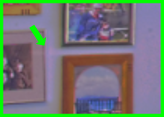}
			\end{minipage}
		\end{minipage}
		\label{fig11a}}
	\subfloat[\small GFES(SP-2)]{
		\begin{minipage}[b]{0.45\linewidth}
			\includegraphics[width=1\linewidth]{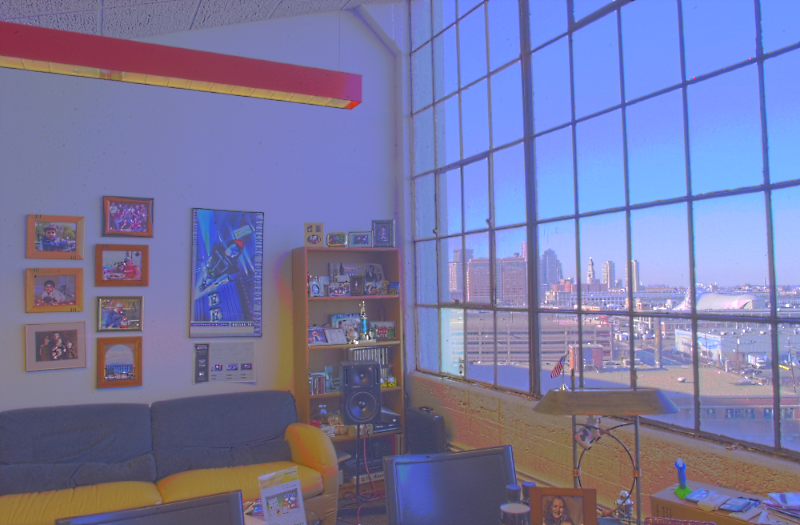}
			\begin{minipage}[b]{0.485\linewidth}
				\includegraphics[width=1\linewidth]{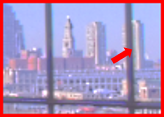}
			\end{minipage}
			\begin{minipage}[b]{0.485\linewidth}
				\includegraphics[width=1\linewidth]{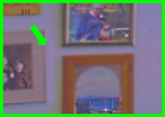}
			\end{minipage}
		\end{minipage}
		\label{fig11b}}
	\hfil
	\subfloat[\small ILS]{
		\begin{minipage}[b]{0.45\linewidth}
			\includegraphics[width=1\linewidth]{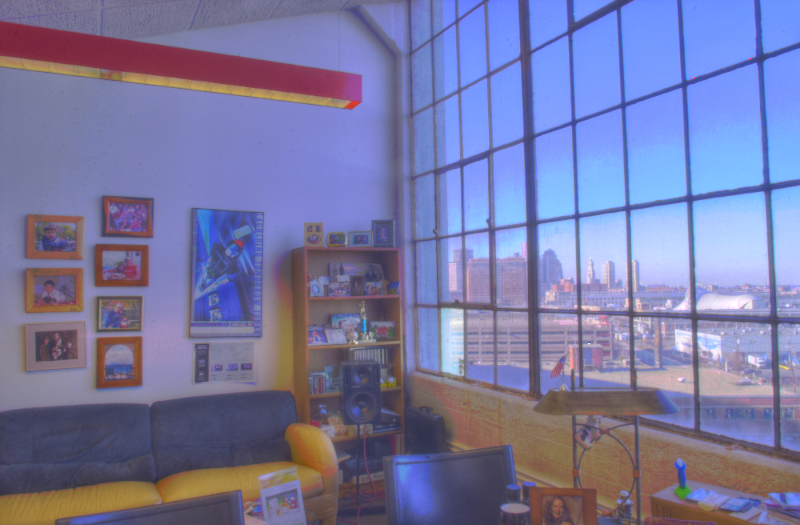}
			\begin{minipage}[b]{0.485\linewidth}
				\includegraphics[width=1\linewidth]{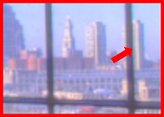}
			\end{minipage}
			\begin{minipage}[b]{0.485\linewidth}
				\includegraphics[width=1\linewidth]{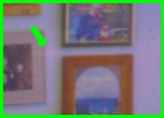}
			\end{minipage}
		\end{minipage}
		\label{fig11c}}
	\subfloat[\small Ours]{
		\begin{minipage}[b]{0.45\linewidth}
			\includegraphics[width=1\linewidth]{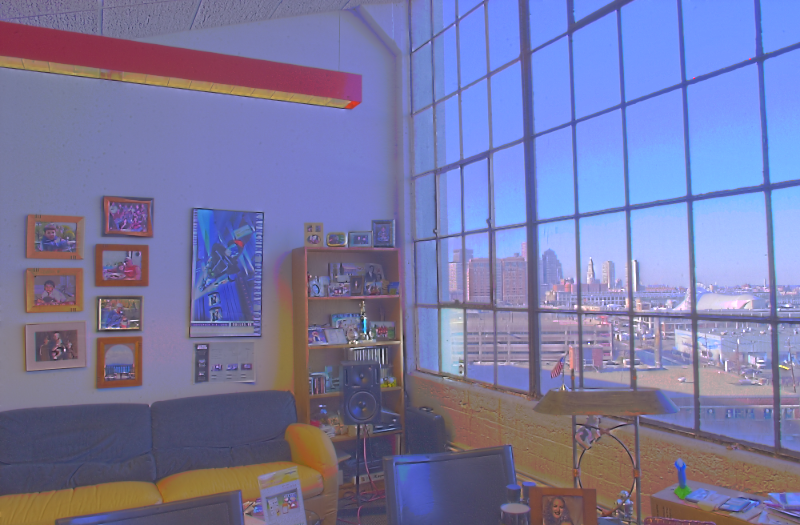}
			\begin{minipage}[b]{0.485\linewidth}
				\includegraphics[width=1\linewidth]{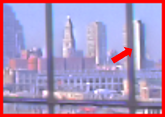}
			\end{minipage}
			\begin{minipage}[b]{0.485\linewidth}
				\includegraphics[width=1\linewidth]{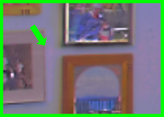}
			\end{minipage}
		\end{minipage}
		\label{fig11d}}
	\caption{HDR tone mapping results by (a) GFES of the EP-1 mode ($\lambda $ = 10, $\alpha$ = 1.2), (b) GFES of the SP-2 mode ($\lambda$ = 100), (c) ILS ($p$ = 1, $\lambda$ = 10) and (d) Ours ($\lambda$ = 0.1, $N$ = 2, $\sigma$ = 2). Clearly, we can enhance the details better than the others, as shown in the red boxes, while alleviating the blocky artifacts, as marked by the arrows in the green boxes. (Zoom in for a better view.)}
	\label{fig11}
\end{figure}

\begin{figure}[htb]
	\centering
	\subfloat[\small Input]{
		\begin{minipage}[b]{0.31\linewidth}
			\includegraphics[width=1\linewidth]{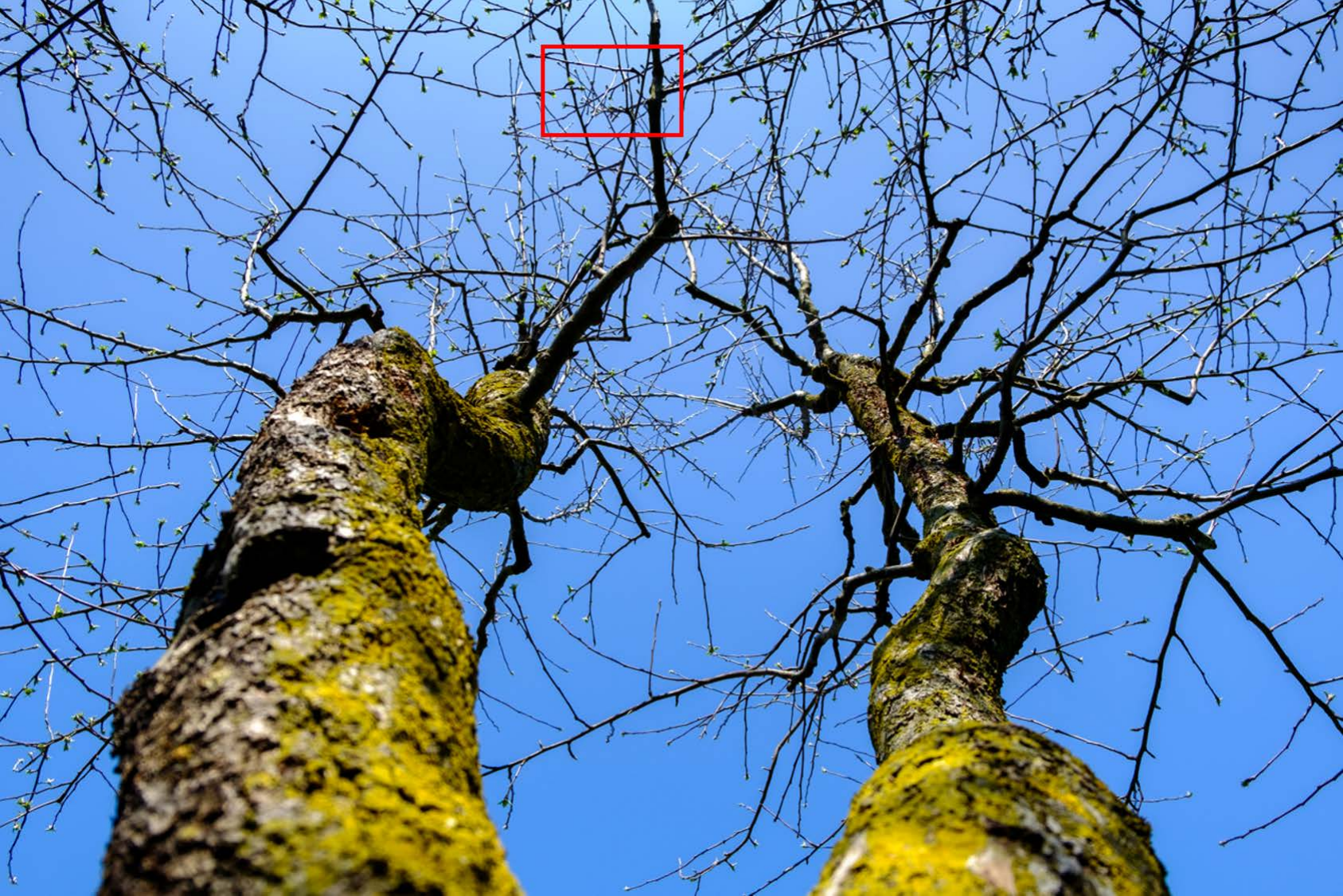}
			\begin{minipage}[b]{0.48\linewidth}
				\includegraphics[width=1\linewidth]{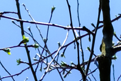}
			\end{minipage}
			\begin{minipage}[b]{0.48\linewidth}
				\includegraphics[width=1\linewidth]{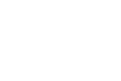}
			\end{minipage}
		\end{minipage}
		\label{fig12a}}	
	\subfloat[\small ILS]{
		\begin{minipage}[b]{0.31\linewidth}
			\includegraphics[width=1\linewidth]{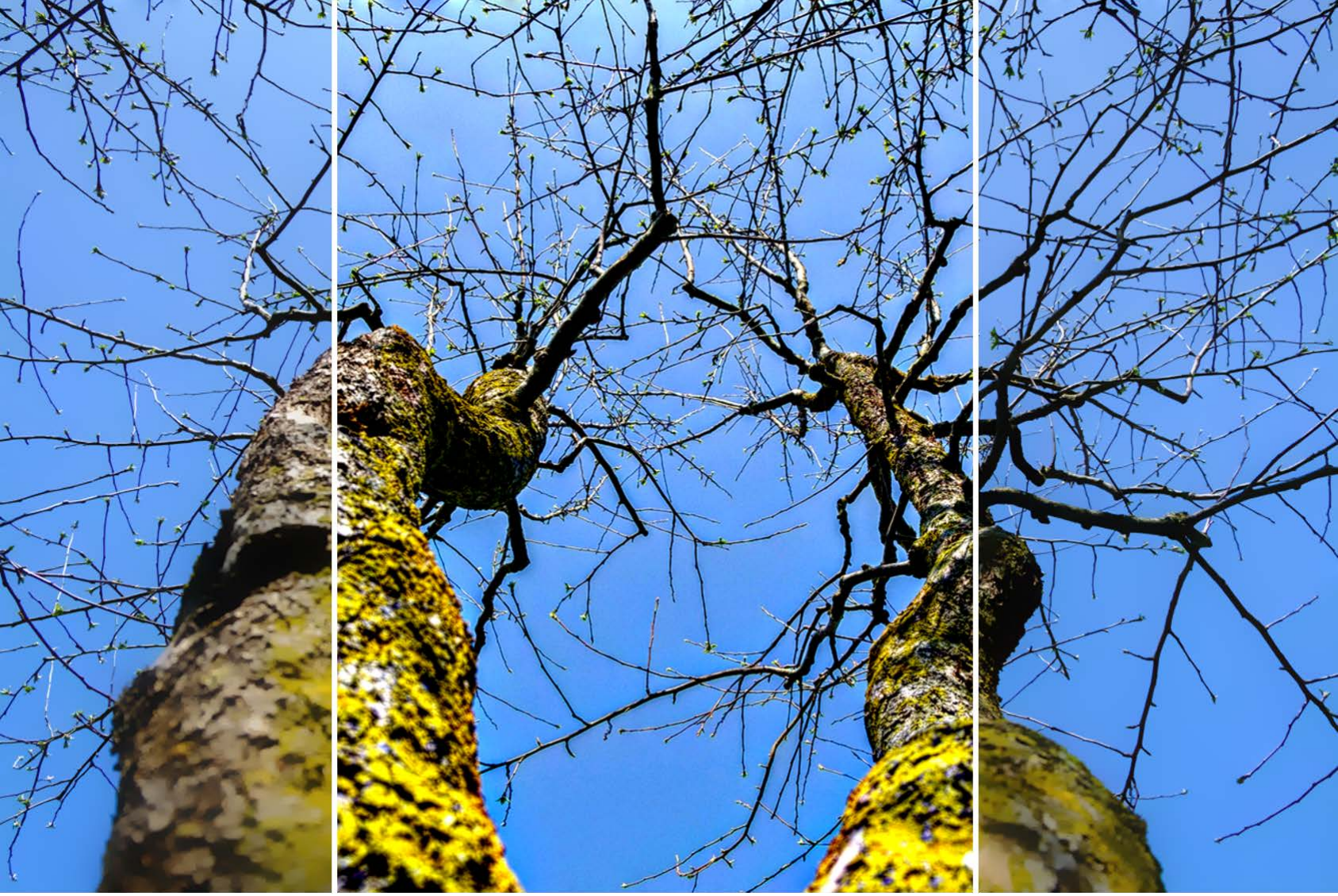}
			\begin{minipage}[b]{0.48\linewidth}
				\includegraphics[width=1\linewidth]{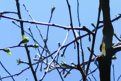}
			\end{minipage}
			\begin{minipage}[b]{0.48\linewidth}
				\includegraphics[width=1\linewidth]{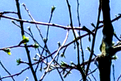}
			\end{minipage}
		\end{minipage}
		\label{fig12b}}
	\subfloat[\small Ours]{
		\begin{minipage}[b]{0.31\linewidth}
			\includegraphics[width=1\linewidth]{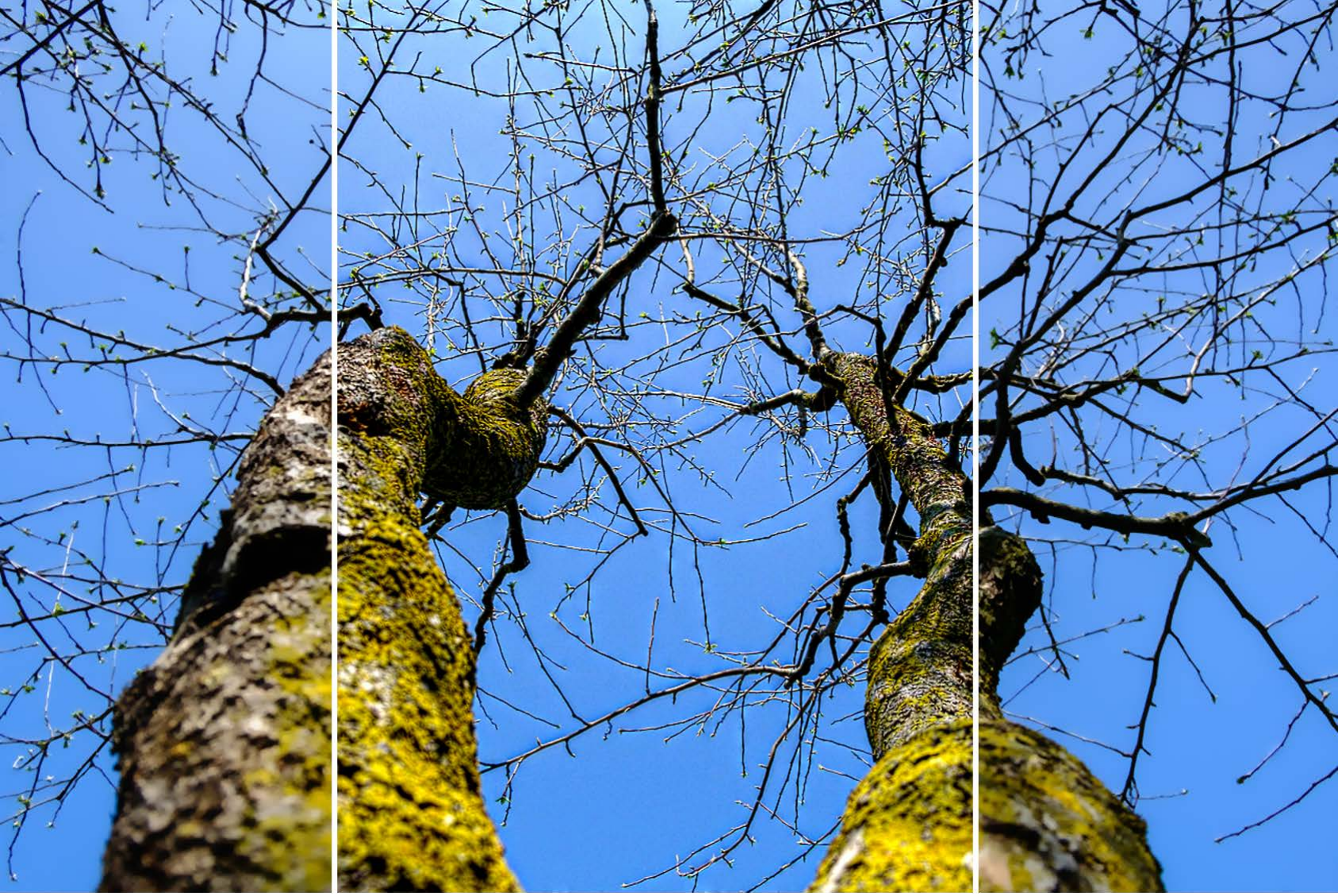}
			\begin{minipage}[b]{0.48\linewidth}
				\includegraphics[width=1\linewidth]{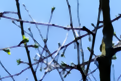}
			\end{minipage}
			\begin{minipage}[b]{0.48\linewidth}
				\includegraphics[width=1\linewidth]{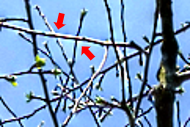}
			\end{minipage}
		\end{minipage}
		\label{fig12c}}
	\caption{ Comparison about compartmentalization artifacts in image detail enhancement. The smoothed results are produced by (b) ILS ($\lambda$ = 1.0, $N$ = 4) and (c) Ours ($\lambda$ = 0.1, $N$ = 2, $\sigma$ = 1). Top: the left and right parts in (b) and (c) are the smoothed images and the middle parts are the corresponding 3× detail enhanced images. Bottom: Close-up of the red box in (a) and the corresponding smoothed results (the left) and enhanced results (the right). Compared to ILS, our method produces a little more evident compartmentalization artifacts, as marked by the red arrows in the enlarged box in (c).}
	\label{fig12}
\end{figure}
\section{Conclusions}
\label{sec:conclu}

It is still challenging with existing methods to smooth out high-contrast details while preserving weak structures, as they are difficult to distinguish structures and details to handle them distinctively, due to the overlapped ranges of gradients for structures and details. In this paper, we address this challenge by developing novel measures to determine smoothing manners for pixels via their properties on structures or not, no matter whether they have high or low gradients. In this way, we can well distinguish structures and details to handle distinctively, and so convenient for preserving weak structures while smoothing out high-contrast details. Moreover, we can still use the framework of ILS for real time image smoothing while using fewer iterations than the original ILS for acceleration. In sum, we present a novel image smoothing method that can more efficiently produce better results than existing methods, especially on preserving weak structures while removing high-contrast details.

{
\small
\bibliographystyle{ieee_fullname}
\bibliography{egbib}
}


\end{document}